\newif\ifreview 
\newif\ifarxiv \newcommand{\arxiv}{\arxivtrue}
\newif\ifcamera 

\arxiv

\documentclass[11pt]{article}

\ifreview \usepackage[review]{acl} \fi
\ifarxiv \usepackage[final]{acl} \fi
\ifcamera \usepackage[final]{acl} \fi

\usepackage[utf8]{inputenc}
\usepackage{lmodern}
\usepackage{times}
\usepackage{latexsym}
\usepackage[T1]{fontenc}

\usepackage{microtype}

\usepackage{inconsolata}

\usepackage{bbding}
\usepackage{graphicx}
\usepackage{amsmath}
\usepackage{amssymb}
\usepackage{booktabs}
\usepackage{amsfonts}
\usepackage{algorithm}
\usepackage{algorithmicx}
\usepackage{algpseudocode}
\usepackage{footnote}

\usepackage{fancybox}
\usepackage{tcolorbox}
\usepackage{enumitem}

\usepackage{tikz}
\usetikzlibrary{shapes, arrows, positioning}
\usepackage{xcolor}
\usepackage{comment}
\usepackage{amsthm}
\usepackage{bbm} 
\usepackage{subcaption}
\usepackage{multirow}
\usepackage{comment}
\usepackage{makecell} 
\usepackage{tabularx}
\usepackage[normalem]{ulem}
\useunder{\uline}{\ul}{}
\usepackage{booktabs}
\usepackage[switch]{lineno}

\usepackage{url}
\usepackage{xspace}
\usepackage[export]{adjustbox}

\usepackage{color}
\usepackage{colortbl}

\newcommand{\nbf}[1]{{\noindent \textbf{#1}}}

\usepackage[capitalize]{cleveref}
\crefname{appendix}{App.}{App.}
\crefname{section}{Sec.}{Secs.}
\crefname{table}{Table}{Tables}
\crefname{figure}{Fig.}{Figs.}
\crefname{algocf}{alg.}{algs.}
\Crefname{algocf}{Algorithm}{Algorithms}

\definecolor{promptbg}{HTML}{DEEBF7}
\definecolor{promptframe}{HTML}{3182BD}

\usepackage[utf8]{inputenc}

\def\paperTitle{SIN-Bench: Tracing Native Evidence Chains in Long-Context Multimodal Scientific Interleaved Literature}

\def\authorBlock{
    Yiming Ren$^\textnormal{1,2}$\footnotemark[1] \qquad
    Junjie Wang$^\textnormal{1,3}$\footnotemark[1] \qquad
    Yuxin Meng$^\textnormal{1}$\footnotemark[1] \qquad
    Yihang Shi$^\textnormal{1}$\footnotemark[1]
    \\
    \textbf{
    Zhiqiang Lin$^\textnormal{1}$ \qquad
    Ruihang Chu$^\textnormal{1}$ \qquad
    Yiran Xu$^\textnormal{1}$  \qquad
    Ziming Li$^\textnormal{4}$  \qquad
    Yunfei Zhao$^\textnormal{3,5}$
    } \\
    \textbf{
    Zihan Wang$^\textnormal{3,6}$ \qquad
    Yu Qiao$^\textnormal{2}$\footnotemark[2]  \qquad
    Ruiming Tang$^\textnormal{4}$ \qquad
    Minghao Liu$^\textnormal{3}$ \qquad
    Yujiu Yang$^\textnormal{1}$\footnotemark[2]
    }
    \vspace{-0.25em}
    \\\\ 
    \selectfont{$^\textnormal{1}$Tsinghua University} \quad
    \selectfont{$^\textnormal{2}$Shanghai AI Laboratory} \quad
    \selectfont{$^\textnormal{3}$2077AI} \\
    \selectfont{$^\textnormal{4}$KuaiShou Inc.} \quad
    \selectfont{$^\textnormal{5}$Stanford University} \quad
    \selectfont{$^\textnormal{6}$Harvard University}
    \\
    {\tt\small rym24@mails.tsinghua.edu.cn}, \quad
    {\tt\small wangjunjie@sz.tsinghua.edu.cn}, \quad
    {\tt\small qiaoyu@pjlab.org.cn}, \quad
    {\tt\small yang.yujiu@sz.tsinghua.edu.cn}
    \\
    \url{https://github.com/IIGROUP/sin-bench}
}

\title{\paperTitle}

\author{\authorBlock}

\begin{document}

\maketitle

{
  \renewcommand{\thefootnote}%
  {\fnsymbol{footnote}}
  \footnotetext[1]{Equal contribution.}
  \footnotetext[2]{Corresponding Author.}
  \footnotetext{Under Review.}
}

\newcommand{\data}{SIN-Data}  
\newcommand{\bench}{SIN-Bench}  

\setlength{\textfloatsep}{13pt}
\setlength{\floatsep}{6pt}

\begin{abstract}
  
Evaluating whether multimodal large language models truly understand long-form scientific papers remains challenging: 
answer-only metrics and synthetic ``Needle-In-A-Haystack'' tests often reward answer matching without requiring a causal, evidence-linked reasoning trace in the document.
We propose the ``Fish-in-the-Ocean'' (FITO) paradigm, which requires models to construct explicit cross-modal evidence chains within native scientific documents.
To operationalize FITO, we build \texttt{SIN-Data}, a scientific interleaved corpus that preserves the native interleaving of text and figures.
On top of it, we construct \texttt{SIN-Bench} with four progressive tasks covering evidence discovery (SIN-Find), hypothesis verification (SIN-Verify), grounded QA (SIN-QA), and evidence-anchored synthesis (SIN-Summary).
We further introduce ``No Evidence, No Score'', scoring predictions when grounded to verifiable anchors and diagnosing evidence quality via matching, relevance, and logic.
Experiments on eight MLLMs show that grounding is the primary bottleneck: Gemini-3-pro achieves the best average overall score ($0.566$), while GPT-5 attains the highest SIN-QA answer accuracy ($0.767$) but underperforms on evidence-aligned overall scores, exposing a gap between correctness and traceable support.

\end{abstract}

\section{Introduction}
The deep analysis of long-form, multimodal scientific literature represents a hallmark of human cognition~\citep{DBLP:journals/corr/abs-2406-11633@DocGenome,DBLP:journals/chinaf/ChenWTYGCTHLMMWDYGHSJXW24@how-far,DBLP:journals/corr/abs-2508-15763@Intern-S1}. 
Accordingly, this task serves as a vital benchmark for the assessment of reasoning capabilities at the expert level, such as the proficiency of doctoral researchers. 
Advancements in Multimodal Large Language Models (MLLMs)~\citep{@Qwen3-VL,openai2025@gpt-5,google2025@gemini-3-pro,xai2025@grok-4}, particularly in long-context processing and multimodal alignment, render the attainment of this expert reading ability achievable. 
However, when models process these lengthy and symbol-dense documents, existing evaluation systems fail to distinguish \textit{whether the system genuinely comprehends the complex document logic or merely relies on parametric knowledge to infer the answer}.

\begin{figure}[t] 
\centering
\includegraphics[width=0.5\textwidth]{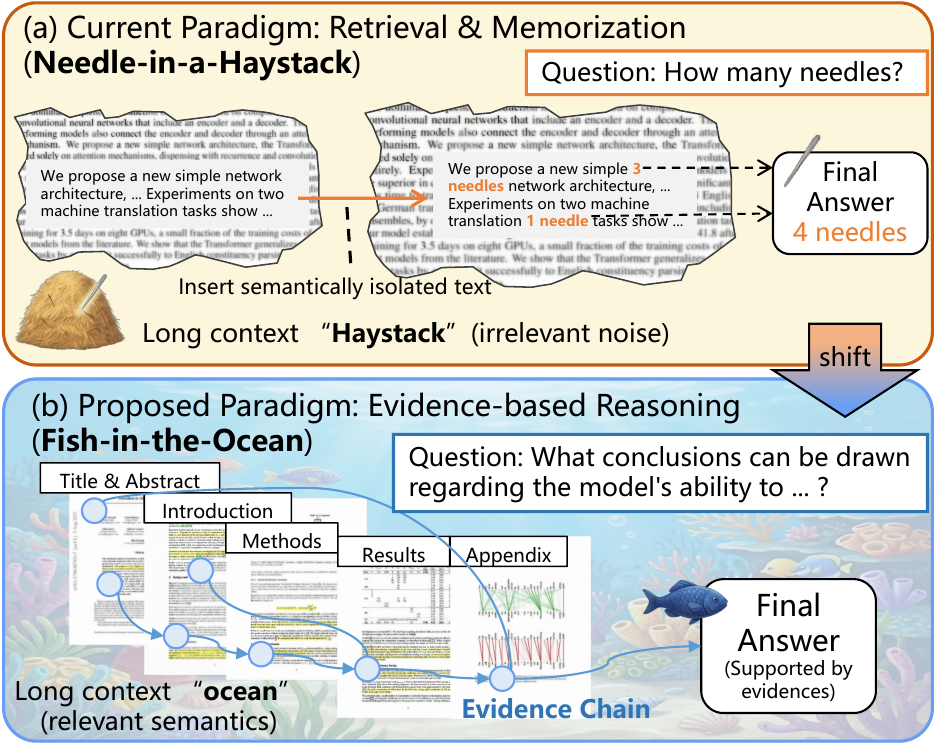} 
\caption{
Comparison of long-context multimodal evaluation paradigms. 
(a) Current NIAH approaches embed artificial ``needles'' into irrelevant noise, focusing on surface-level retrieval. 
(b) The proposed FITO paradigm evaluates deep comprehension within the native document ecosystem (``ocean''). 
It requires the model to aggregate interconnected knowledge units (``fish'') across sections to form an evidence chain for reasoning.
}
\label{fig:teaser} 
\end{figure}

Despite the rapid progress in long-context evaluation, a significant misalignment persists between existing benchmarks and authentic scientific cognition~\citep{DBLP:journals/corr/abs-2411-15296@mme-survey}. 
This discrepancy manifests primarily in task construction and evaluation metrics. 
First, for \textbf{task design}, as shown in~\cref{fig:teaser} (a), the prevailing ``Needle In A Haystack'' (NIAH) paradigm favors the insertion of artificially constructed and semantically isolated segments~\citep{DBLP:conf/nips/WangZRDLLH0ZLZL24@mm-niah}. 
While this synthetic configuration effectively tests retrieval boundaries, it fails to simulate the nativeness and long-range dependencies inherent in genuine literature. 
For instance, the interpretation of figures frequently demands a retrospective reference to experimental setups within the methodology sections. 
Second, concerning \textbf{evaluation metrics}, an exclusive reliance on answer accuracy obscures the opacity of the reasoning process~\citep{DBLP:conf/wacv/MathewKJ21@DocVQA,DBLP:journals/corr/abs-2406-11633@DocGenome}. 
This approach allows models to exploit parametric knowledge as a shortcut, which causes hallucinations when the system confronts scientific inquiries.

To bridge this gap, we advocate a fundamental shift in evaluation focus from mere correctness of answers to the construction of cross-modal evidence chains.
Accordingly, as shown in~\cref{fig:teaser} (b), we introduce the ``Fish-in-the-Ocean'' (FITO) evaluation paradigm. 
To demonstrate that the model avoids mere speculation, the system must explicate the process of evidence retrieval~\citep{DBLP:conf/naacl/DasigiLBCSG21@qasper}. 
Distinct from the artificially implanted ``needle'' (artificial facts within noise) characteristic of NIAH, critical information within ``ocean'' (scientific literature) resembles a native ``fish'' (interconnected knowledge). 
Consequently, the resolution of scientific inquiries essentially requires the capture and correlation of these ``fish'' dispersed across components such as the abstract, figures, and appendices. 
The core competency of a model lies in the identification of high-value evidence from the vast information space and the aggregation of these findings into a self-consistent logical chain. 

To operationalize the Fish-in-the-Ocean (FITO) paradigm, we build \texttt{SIN-Data}, a unified \textbf{S}cientific \textbf{IN}terleaved data infrastructure tailored for long-form scientific documents.
We aggregate papers from sources such as arXiv and PMC, and parse each document into a unified representation that preserves the native alternation of prose and visual evidence (e.g., figures) in its original reading order.
Following the arXiv taxonomy, \texttt{SIN-Data} covers more than $12$ primary disciplines and over $80$ subfields, yielding a broad substrate of domain knowledge.
Building on \texttt{SIN-Data}, we develop a scalable pipeline to construct \texttt{SIN-Bench}.
Each instance is defined by a query, answer(s), and an evidence chain grounded to native anchors.
As a hierarchical evaluation suite aligned with real scientific cognition, \texttt{SIN-Bench} instantiates a progressive ``discovery--verification--synthesis'' workflow via four tasks:
(1) SIN-Find (evidence discovery),
(2) SIN-Verify (hypothesis verification),
(3) SIN-QA (grounded reasoning), and
(4) SIN-Summary (evidence-anchored synthesis).
To avoid isolated, answer-only evaluation, we adopt the principle of \textbf{``No Evidence, No Score''} principle:
High-confidence reasoning is credited only when it is grounded in verifiable evidence.
Beyond answer accuracy, we assess evidence quality along three dimensions: Logic, Matching, and Relevance.

Empirically, evaluating eight widely-used MLLMs reveals that evidence grounding is the primary bottleneck in long-form scientific reading.
For example, Gemini-3-pro achieves the best average overall score ($0.566$), while GPT-5 attains the highest SIN-QA answer accuracy ($0.767$) yet lags in evidence-aligned overall scores on SIN-Find and SIN-QA, indicating that correct answers do not necessarily imply traceable support.
Furthermore, structured evidence output remains challenging for several open-weight models (often yielding invalid scores).

In summary, the contributions are as follows:
\begin{itemize}[nosep, leftmargin=*]
    \item We introduce the ``Fish-in-the-Ocean'' paradigm for evaluating evidence-based long-context multimodal scientific reasoning.
    \item We build \texttt{SIN-Data} and \texttt{SIN-Bench}, a four-task suite spanning discovery, verification, QA, and evidence-anchored synthesis.
    \item We propose ``No Evidence, No Score'' and multi-dimensional evidence metrics to diagnose grounding quality beyond answer accuracy.
\end{itemize}

\section{Related Work}

\nbf{MLLM Long-Context Understanding Tasks.}
Recent progress in long-context understanding for MLLMs primarily focuses on exploring retrieval boundaries. 
Early paradigms, such as MM-NIAH~\citep{DBLP:conf/nips/WangZRDLLH0ZLZL24@mm-niah} and MMLongCite~\citep{DBLP:journals/corr/abs-2510-13276@MMLongCite}, adopt a Needle-In-A-Haystack setup that inserts synthetic text or image needles into long contexts to test recall limits.
These methods effectively quantify context capacity, but they rely on semantically isolated synthetic content and do not reflect the complex logical dependencies in real cognition.
To improve realism, later benchmarks such as MMLongBench-Doc~\citep{DBLP:conf/nips/MaZC0JLLLMDZP0W24@mmlongbench-doc} and LongDocURL~\citep{DBLP:conf/acl/DengYBWLXLGS0025@LongDocURL} move to real web pages and multi-page reports. 
However, these benchmarks mainly emphasize single-point information extraction or page layout perception, and they often overlook long-range cross-modal reasoning that spans the document. 
To address this gap, we propose the Fish-In-The-Ocean paradigm. 
Unlike isolated retrieval tasks, this paradigm requires the construction of a coherent cross-modal evidence chain. 
This work further introduces \texttt{SIN-Bench}, which simulates a complete scientific research workflow and evaluates deep reasoning in highly interconnected scientific literature.

\nbf{Evaluating Scientific Reasoning in MLLMs.}
Scientific literature contains dense multimodal content and therefore serves as an effective domain for evaluating expert-level reasoning. 
Several efforts focus on converting PDF and LaTex documents into machine-readable formats and provide a foundation for subsequent research~\citep{DBLP:journals/corr/abs-2406-13923@pin,DBLP:conf/icdar/ZhongTJ19@PubLayNet,DBLP:journals/corr/abs-2406-11633@DocGenome}. 
Building on this foundation, recent studies shift toward higher-level cognition. 
M-DocSum~\citep{DBLP:journals/corr/abs-2503-21839@M-DocSum} introduces reference-grounded interleaved text-and-image summarization, and MMIE~\citep{DBLP:conf/iclr/XiaHQZWZCCDLWY25@MMIE} designs information extraction tasks for knowledge-intensive domains. 
However, many benchmarks rely on prediction-oriented metrics, such as ROUGE or simple question answering accuracy, and they do not assess the transparency of the reasoning process. 
This black-box evaluation setting allows answers that come from parametric memorization rather than document-level understanding. 
Inspired by recent works on deep research~\citep{DBLP:journals/corr/abs-2409-12959@MMSearch,DBLP:journals/corr/abs-2506-02454@DeepResearcher}, we introduce an evidence-centered metric suite that includes match, validity, and logic, and it enforces a ``No Evidence, No Score'' principle to enable fine-grained diagnosis of reasoning behavior.

\section{The Fish-in-the-Ocean Paradigm}

Contrary to the ``Needle-in-a-Haystack''~\citep{DBLP:conf/nips/WangZRDLLH0ZLZL24@mm-niah} paradigm, which relies on inserting synthetic noise, the Fish-in-the-Ocean (FITO) paradigm evaluates the comprehension of inherently complex, long-form multimodal documents. 
In this analogy, long-context multimodal documents serve as the canonical ``ocean,'' characterized by three essential properties that challenge expert-level reasoning:
(i) \textbf{Nativeness}: Information is inherently present and semantically entangled, rather than artificially injected;
(ii) \textbf{Interconnectivity}: Evidence spans sections/modalities (e.g., text, figures, tables);
(iii) \textbf{Long-range Dependency}: Logical chains span disjoint sections, such as linking the method to the result and the conclusion.

To evaluate this process, we move beyond simple QA likelihood maximization, $P(A|D, Q)$, where $A$, $D$, and $Q$ denote the answer, document, and query, respectively. 
Instead, FITO models the joint probability of the answer and its supporting evidence-chain ($E$):
\begin{equation}
\small
P(A, E | D, Q) = P(E | D, Q) \cdot P(A | E, D, Q).
\label{eq:fito_prob}
\end{equation}
This formulation shifts the evaluation objective from result-oriented prediction to process-oriented reasoning.
It implies that a valid system must explicitly instantiate the latent variable $E$ and validate its sufficiency before deriving $A$.

\begin{figure*}[!t] 
\centering
\includegraphics[width=1\textwidth]{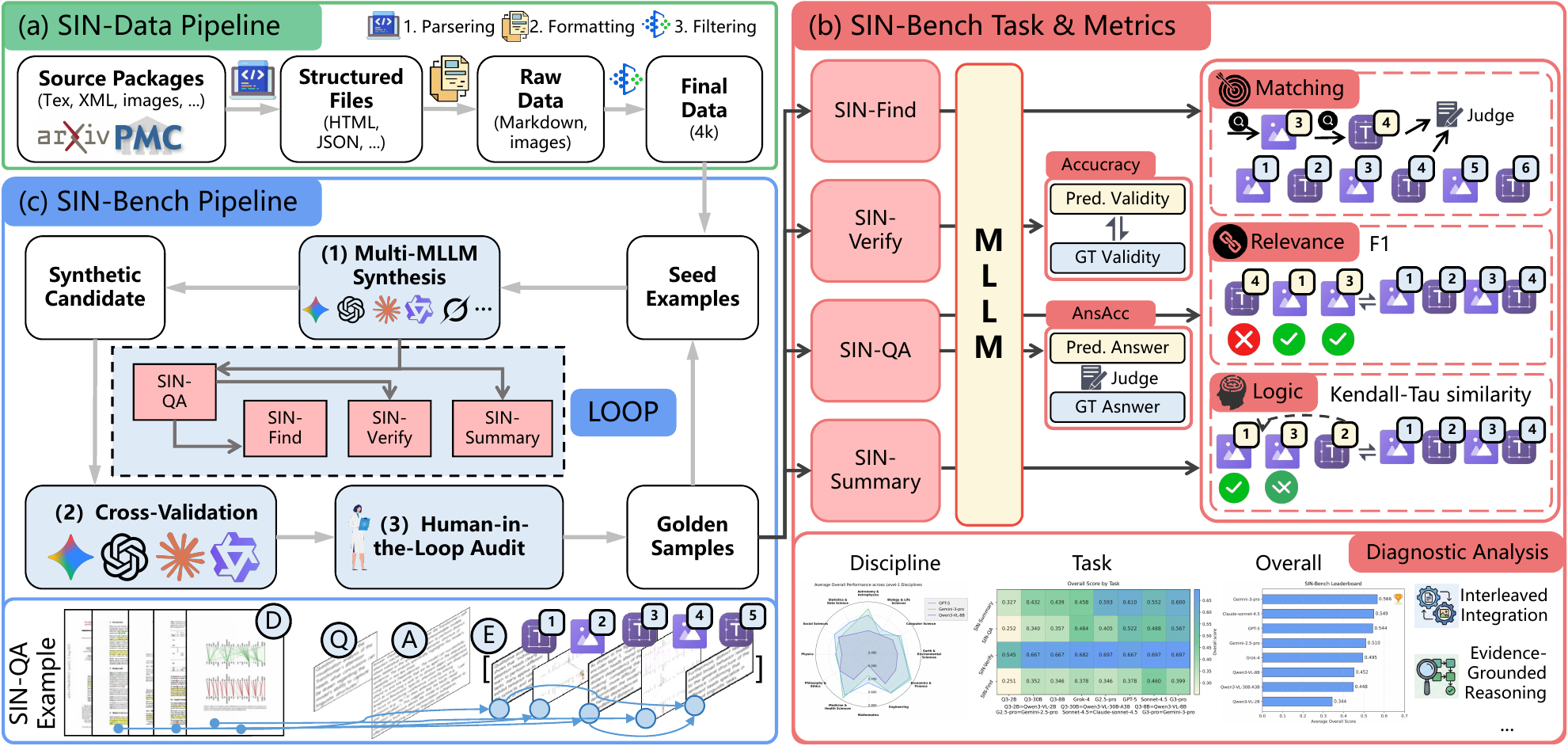} 
\caption{
The framework of \texttt{SIN-Bench}. 
\textbf{(a) SIN-Data Infrastructure}: We parse raw source packages into a unified Scientific Interleaved format using a semantic-first strategy.
\textbf{(c) Construction Pipeline}: Based on the data, we employ an iterative Multi-MLLM synthesis loop with cross-validation and human auditing to generate high-quality samples.
\textbf{(b) Task \& Metrics}: The benchmark features four hierarchical tasks evaluated under the ``No Evidence, No Score'' protocol. We assess evidence chains across Matching, Relevance, and Logic dimensions.
}
\label{fig:sin-data-bench} 
\end{figure*}

\section{SIN-Bench: Evidence-based Evaluation}

To operationalize the FITO paradigm, we introduce \texttt{SIN-Bench}. 
We select scientific literature as the evaluation substrate, as its inherent information density and rigorous logical for assessing expert-level cognition. 
Specifically, we establish a semi-automated framework that transforms raw scientific corpora into structured reasoning challenges. 
As shown in~\cref{fig:sin-data-bench}, this framework encompasses the construction of a unified Scientific INterleaved (SIN) representation (\cref{ss:sin-data}), an evidence-based hierarchical task design (\cref{ss:sin-bench-task}), an evidence-driven metric system (\cref{ss:sin-bench-metrics}), and a scalable benchmark construction pipeline (\cref{ss:sin-bench-pipeline}).

\subsection{SIN-Data Infrastructure}
\label{ss:sin-data}

To facilitate rigorous evaluation of long-context multimodal reasoning, we establish the \texttt{SIN-Data} infrastructure, a pipeline designed to unify original documents into a semantics-aware Scientific INterleaved (SIN) format.
Our primary objective is to transcend the limitations of raw formats (e.g., PDF layout or raw LaTeX) by reconstructing a linear data stream that preserves the logical and, rather than spatial, coupling between text and visual evidence.
As shown in~\cref{fig:sin-data-bench} (a), inspired by recent work~\citep{DBLP:journals/corr/abs-2406-13923@pin}, our pipeline processes $50$k raw source packages from arXiv\footnote{\url{https://arxiv.org/}} and PubMed Central (PMC)\footnote{\url{https://www.ncbi.nlm.nih.gov/pmc/}} through three progressive stages:

\nbf{Stage 1: Element Parsing.} 
We decouple content extraction from presentation layers. 
For arXiv, the source LaTeX is compiled into responsive HTML via Engrafo, followed by text parsing. 
Images are recovered from the DOM tree and re-anchored to their textual context via visual matching. 
For PMC, we parse JATS XML files into structured JSON with the s2orc-doc2json parser, which preserves tables and citation links while stripping decorative artifacts.

\nbf{Stage 2: Semantic-First Formatting.} 
Structured data is unified into Interleaved Markdown. 
Unlike layout-dependent PDF extraction, we propose a Citation-Driven Injection Strategy to preserve the logical ``chain of evidence.'' 
Each visual component is assigned a unique ID $\langle x_k \rangle$ and inserted near the paragraph of its first citation (e.g., before ``Figure 3 shows that...''). 
This ensures visual evidence aligns with the reasoning logic.
We concurrently compute quality signals, such as the count of text-image interleaved segments and token density, for subsequent processing.

\nbf{Stage 3: Quality Filtering.}
To ensure high multimodal density, we rigorously prune samples with sparse visual context or broken citation links by quality signals. 
Moreover, we reference the official arXiv taxonomy to categorize and annotate samples, ensuring the retention of a representative distribution across diverse domains.

This process yields a final set of $4,000$ high-quality documents ($D$). 
These samples cover a diverse range of $12$ top-level disciplines and over $80$ subfields. 
Details are in~\cref{ass:sin-data-pipeline}.

\subsection{Evidence-based Hierarchical Tasks}
\label{ss:sin-bench-task}

To operationalize the evidence-chain-driven objective of the FITO paradigm into actionable benchmark instances, \texttt{SIN-Bench} abstracts the professional reading workflow into four hierarchical tasks: Discovery $\rightarrow$ Verification $\rightarrow$ Question Answering (QA) $\rightarrow$ Synthesis. 
Each instance employs a single document $D$ as the substrate, organizing annotations around a query $Q$, an answer $A$, and an evidence chain $E = [e_1, \dots, e_N]$, structured $N$ interleaved anchors, where odd-indexed elements denote visual anchors and even-indexed elements represent text spans.
This unified interface ensures semantic alignment across all four tasks, facilitating consistent diagnosis.

\nbf{SIN-Find (Evidence Discovery).} 
This task assesses the model's ability to identify reasoning paths within long-context documents. 
Given a document $D$ and a query $Q$, the model must construct the evidence chain $E$:
\begin{equation}
\small
E = f_{\text{find}}(D, Q).
\end{equation}

Unlike keyword-matching retrieval, $Q$ is designed to require complex reasoning (e.g., factor induction or method comparison). 
Consequently, the model must retrieve semantically sufficient cross-sectional or cross-modal information and organize it into a logically ordered chain $E$, rather than merely extracting disjointed snippets.

\nbf{SIN-Verify (Hypothesis Verification).} 
This task targets the auditing capability, determining whether the provided evidence suffices to support a specific conclusion. 
The formulation is a binary classification problem:
\begin{equation}
\small
y = f_{\text{verify}}(D, Q, A, E), \quad y \in \{0, 1\}
\end{equation}
where the model judges if $E$ constitutes consistent and sufficient support for $A$. 
To enforce strict discrimination, we introduce negative samples through systematic perturbations (e.g., omitting premises or mismatching conditions). 
This simulates the critical review process in authentic research environments.

\nbf{SIN-QA (Grounded Reasoning).} 
This task demands the joint generation of the answer and its provenance. 
The model is required to output both the answer $A$ and the evidence chain $E$:
\begin{equation}
\small
(A, E) = f_{\text{qa}}(D, Q).
\end{equation}

Here, $A$ and the explanatory text within $E$ are generated de novo rather than extracted. 
This configuration constrains the model's reasoning to verifiable evidence, effectively distinguishing evidence-driven comprehension from hallucinations based on parametric priors.

\nbf{SIN-Summary (Evidence-Anchored Synthesis).} 
This task evaluates holistic understanding and long-range integration over the full document.
Given $D$, the model generates an interleaved, evidence-anchored summary consisting of multiple claims, each paired with its evidence:
\begin{equation}
\small
\hat{S} = \bigl\{a_j, E_j\bigr\}_{j=1}^{J} = f_{\text{sum}}(D).
\label{eq:sum}
\end{equation}

Under our interleaved response format, the overall ``answer'' and ``evidence'' are overlapping: the synthesis can be equivalently viewed as
\begin{equation}
\small
A=\{a_j\}_{j=1}^{J}, \qquad 
E=\bigcup_{j=1}^{J}E_j,
\label{eq:sum_overlap}
\end{equation}
where each statement $a_j$ is accompanied by anchors that make it verifiable.
By requiring both cross-sectional integration and explicit evidence for each claim, this task operationalizes the long-range dependencies and multimodal connectivity.

\subsection{Metrics: No Evidence, No Score}
\label{ss:sin-bench-metrics}

To mitigate evaluation bias where models derive correct answers through incorrect reasoning paths (the ``right for the wrong reasons'' phenomenon), \texttt{SIN-Bench} adheres to a ``No Evidence, No Score'' philosophy. 
As shown in~\cref{fig:sin-data-bench} (b), we structure the evaluation framework into two layers. 
For tasks that require generating evidence chains (SIN-Find, SIN-QA, and SIN-Summary), we employ a shared protocol to assess the quality of the evidence. 
Building on this foundation, SIN-QA incorporates an additional assessment of the answer, while SIN-Verify relies on accuracy.

\nbf{Interleaved Evidence Chain.}
We denote the ground-truth evidence chain as $E^\star=[e^\star_1,\dots,e^\star_N]$ and the predicted chain as $\hat{E}=[\hat{e}_1,\dots,\hat{e}_M]$. 
Consistent with the unified interface, elements at odd indices represent visual anchors, and elements at even indices correspond to text spans. 
We define the grouping components as:
\begin{equation}
\small
\begin{aligned}
    K^\star &= \lfloor N/2 \rfloor, & \hat{K} &= \lfloor M/2 \rfloor, \\
    (v^\star_i, t^\star_i) &= (e^\star_{2i-1}, e^\star_{2i}), & (\hat{v}_i, \hat{t}_i) &= (\hat{e}_{2i-1}, \hat{e}_{2i}).
\end{aligned}
\end{equation}

During evaluation, we treat the adjacent pair $(v_i, t_i)$ as the minimal unit of evidence and compute the quality metrics at this granularity.

For the SIN-Find, SIN-QA, and SIN-Summary tasks, we assess the quality of the evidence chain via the MRL (Matching, Relevance, and Logic) metrics. 
We employ a LLM as the evaluator to robustly capture semantic equivalence in long-form evidence, overcoming the rigidity of traditional n-gram metrics. 
Specifically, M involves semantic adjudication by the LLM (e.g., Qwen3~\citep{DBLP:journals/corr/abs-2505-09388@qwen3}; see~\cref{ass:judge-in-metrics} for selection), whereas R and L rely on deterministic analytical formulations.

\nbf{(M) Matching.}
Let $V^\star=\{v^\star_i\}_{i=1}^{K^\star}$ denote the set of ground-truth visual anchors. 
We define the hit set $H$ as the subset of indices in the prediction where the visual anchor matches a ground-truth anchor: $H=\{i\in[1,\hat{K}]\mid \hat{v}_i\in V^\star\}$.
For each index $i\in H$, we identify the corresponding ground-truth index $j(i)$ such that $v^\star_{j(i)}=\hat{v}_i$. 
We then employ the LLM to evaluate the semantic consistency between the predicted text $\hat{t}_i$ and the reference text $t^\star_{j(i)}$, assigning a score $s_i\in\{0,1,2,3\}$. 
We normalize this score as $\tilde{s}_i=s_i/3$. The matching metric is defined as:
\begin{equation} 
\small
\mathrm{Match}= \begin{cases} \frac{1}{|H|}\sum\limits_{i\in H}\tilde{s}_i, & |H|>0,\\ 0, & |H|=0. 
\end{cases} 
\label{eq:match} 
\end{equation}

\nbf{(R) Relevance (F1).}
We formulate the correctness of an evidence unit as a binary classification problem.
The indicator function for a correct prediction is $\mathbb{I}_i=\mathbf{1}[\hat{v}_i\in V^\star \wedge \tilde{s}_i\ge \tau]$, where $\tau$ represents the semantic threshold (set to $2/3$ in our implementation). 
Letting the number of True Positives be $TP=\sum_{i=1}^{\hat{K}}\mathbb{I}_i$, we calculate:
\begin{equation}
\small
\mathrm{Prec}=\frac{TP}{\hat{K}},\quad
\mathrm{Rec}=\frac{TP}{K^\star},\quad
\mathrm{F1}=\frac{2\cdot \mathrm{Prec}\cdot \mathrm{Rec}}{\mathrm{Prec}+\mathrm{Rec}}.
\label{eq:f1}
\end{equation}

\nbf{(L) Logic (Kendall--Tau similarity).}
For the set of matched anchors, let $\hat{\pi}$ and $\pi^\star$ represent their relative permutations in the prediction and the ground truth, respectively, and let $K=|H|$ denote the total number of matches. 
The Kendall--Tau correlation coefficient~\citep{kendall1948rank@kendall} is derived as:
\begin{equation}
\small
\tau=1-\frac{2\cdot \mathrm{inv}(\hat{\pi},\pi^\star)}{K(K-1)/2}\in[-1,1],
\mathrm{KT\text{-}sim}=\frac{\tau+1}{2}\in[0,1].
\label{eq:kt}
\end{equation}

\nbf{SIN-QA: Answer Accuracy.}
For the predicted answer $\hat{A}$, a LLM assigns a semantic correctness score $g \in \{0, 1, 2, 3\}$. 
We normalize this value to derive the answer accuracy metric, $\mathrm{AnsAcc} = g/3$. 

\nbf{SIN-Verify: Accuracy.}
This task functions as a binary classification problem. We evaluate performance using standard accuracy over $T$ samples:
\begin{equation}
\small
\mathrm{Score}_{\textsc{Verify}} = \mathrm{Acc} = \frac{1}{T}\sum{i=1}^{T}\mathbf{1}[\hat{y}_i = y_i].
\label{eq:score_verify}
\end{equation}

Furthermore, we compute the overall score for each task by averaging its metrics.

\subsection{Benchmark Construction Pipeline}
\label{ss:sin-bench-pipeline}

Building upon the interleaved samples provided by \texttt{SIN-Data}, we design an iterative human-model collaborative pipeline. 
It achieves scalability through an iterative synthesis process while ensuring factual correctness, reasoning consistency, and evidence verifiability via cross-model filtering and human-in-the-loop auditing. 
As shown in~\cref{fig:sin-data-bench} (c), the pipeline consists of the following key steps:

\nbf{Seed Examples.} 
We manually create a small set of high-quality seed examples for each task. 

\nbf{(1) Multi-MLLM Synthesis.}
Given a single document $D$, we employ multiple MLLMs for collaborative synthesis. 
We designate SIN-QA and SIN-Summary as the core generation pivot.
The model jointly produces the query $Q$, the answer $A$, and the draft evidence chain $E$ within a single context, thereby minimizing semantic drift in SIN-QA. 
Subsequently, we derive other tasks within the same synthesis round. 
We reformulate SIN-Find by reversing SIN-QA to focus on locating the evidence chain that supports the conclusion. 
For SIN-Verify, we construct negative samples by systematically perturbing $E$ (e.g., through random shuffling) or by creating hard samples that feature insufficient evidence but superficially plausible conclusions. 
For SIN-Summary, models generate a comprehensive summary with ``Cite-as-you-write'' prompting. 
This step yields a pool of candidate samples.

\nbf{(2) Cross-validation.}
To mitigate potential synthesis noise, we implement cross-validation across three powerful models from a model set. 
They independently evaluate the candidate samples on a scale of $1$ to $5$ across three dimensions: rationality of the question, correctness of the answer, uniqueness and consistency of the supporting evidence. 
We strictly retain a sample only if it receives a majority vote and scores $\ge 4$ in all three dimensions. 

\nbf{(3) Human-in-the-Loop Audit.} 
This process focuses on verifying the precise location of evidence anchors and the rigor of the supporting relationships. 
Human experts also correct residual factual inaccuracies and formatting deviations to produce high-confidence golden samples.

\nbf{Iterative Loop.}
These verified golden samples feed back into the system as updated seed examples for subsequent cycles until the dataset reaches the target scale and quality stability.

Utilizing $4,000$ high-quality scientific documents, this iterative process yields approximately $3,200$ raw candidate samples.
After multipe loops, we collect final \texttt{SIN-Bench} with $490$ samples (Find: $159$, QA: $158$, Summary: $89$, Verify: $84$). 
Further details appear in~\cref{ass:sin-bench-pipeline}.

\begin{table*}[t]
\centering
\vspace{-1em}
\resizebox{1\textwidth}{!}{
\renewcommand{\arraystretch}{0.9}
\begin{tabular}{l|cccc|cc|ccccc|cccc|c}
\toprule
\multirow{2.5}{*}{Model}
& \multicolumn{4}{c|}{\textbf{SIN-Find}}
& \multicolumn{2}{c|}{\textbf{SIN-Verify}}
& \multicolumn{5}{c|}{\textbf{SIN-QA}}
& \multicolumn{4}{c|}{\textbf{SIN-Summary}}
& \textbf{Avg.} \\
\cmidrule(lr){2-5} \cmidrule(lr){6-7} \cmidrule(lr){8-12} \cmidrule(lr){13-16}
& L & M & R & \textbf{O}
& Acc & \textbf{O}
& AnsAcc & L & M & R & \textbf{O}
& L & M & R & \textbf{O}
& \textbf{Overall} \\
\midrule
\multicolumn{17}{c}{\textit{Proprietary}} \\ \midrule
Gemini-3-pro
& 0.524 & 0.381 & 0.294 & 0.399
& \textbf{0.697} & \textbf{0.697}
& 0.726 & 0.569 & \textbf{0.521} & \textbf{0.454} & \textbf{0.567}
& 0.784 & \textbf{0.546} & 0.471 & 0.600
& \textbf{0.566} \\
Claude-sonnet-4.5
& 0.512 & \textbf{0.490} & \textbf{0.378} & \textbf{0.460}
& \textbf{0.697} & \textbf{0.697}
& 0.708 & 0.550 & 0.361 & 0.333 & 0.488
& 0.765 & 0.493 & 0.399 & 0.552
& 0.549 \\
GPT-5
& 0.534 & 0.333 & 0.268 & 0.378
& 0.667 & 0.667
& \textbf{0.767} & \textbf{0.578} & 0.406 & 0.336 & 0.522
& \textbf{0.808} & 0.538 & \textbf{0.483} & \textbf{0.610}
& 0.544 \\
Gemini-2.5-pro
& \textbf{0.536} & 0.300 & 0.203 & 0.346
& \textbf{0.697} & \textbf{0.697}
& 0.684 & 0.522 & 0.240 & 0.197 & 0.405
& 0.762 & 0.547 & 0.472 & 0.593
& 0.510 \\
Grok-4
& 0.529 & 0.339 & 0.266 & 0.378
& 0.682 & 0.682
& 0.688 & 0.522 & 0.371 & 0.289 & 0.464
& 0.699 & 0.399 & 0.275 & 0.458 
& 0.495 \\
\midrule
\multicolumn{17}{c}{\textit{Open-weight}} \\
\midrule
Qwen3-VL-8B
& 0.500 & 0.293 & \textbf{0.245} & 0.346
& \textbf{0.667} & \textbf{0.667}
& \textbf{0.476} & \textbf{0.514} & \textbf{0.341} & \textbf{0.234} & \textbf{0.357}
& \textbf{0.683} & \textbf{0.359} & \textbf{0.276} & \textbf{0.439}
& \textbf{0.452} \\
Qwen3-VL-30B-A3B
& \textbf{0.522} & \textbf{0.307} & 0.226 & \textbf{0.352}
& \textbf{0.667} & \textbf{0.667}
& 0.441 & 0.528 & 0.321 & 0.214 & 0.340
& 0.676 & 0.345 & 0.275 & 0.432
& 0.448 \\
Qwen3-VL-2B
& 0.485 & 0.160 & 0.108 & 0.251
& 0.546 & 0.546
& 0.337 & 0.510 & 0.243 & 0.164 & 0.252
& 0.589 & 0.228 & 0.165 & 0.327
& 0.344 \\
\bottomrule
\end{tabular}
}
\caption{Main results on \texttt{SIN-Bench} across diverse MLLMs. We report matching (\textbf{M}), relevance (\textbf{R}), logic (\textbf{L}), answer accuracy (\textbf{AnsAcc}), and verification accuracy (\textbf{Acc}). \textbf{O} denotes the task-specific overall score, and \textbf{Avg. Overall} represents the arithmetic mean of overall scores across all tasks. The best performance is \textbf{bolded}.}
\label{tab:main_results}
\end{table*}

\section{Experiments}
\label{sec:experiments}

\subsection{Experimental Setup}
We examine $8$ widely-used MLLMs, comprising $5$ proprietary and $3$ open-weight models (Details in~\cref{ass:details-baselines}).
Specifically, $5$ proprietary models includes: Gemini-3-pro-preview~\citep{google2025@gemini-3-pro}, Gemini-2.5-pro~\citep{DBLP:journals/corr/abs-2507-06261@gemini-2.5}, GPT-5~\citep{openai2025@gpt-5}, Grok 4~\citep{xai2025@grok-4}, and Claude-sonnet-4.5~\citep{anthropic2025@claude-4.5}.
Considering diversity of architectures in open-weight models, including dense and MoE, we selsect $3$ widely-used models: Qwen3-VL series with Thinking mode (2B, 8B, and 30B-A3B MoE version)~\citep{@Qwen3-VL}.

Furthermore, we present a detailed discussion regarding all employed models and the rationale for the selection in~\cref{ass:model-in-paper}.

\begin{figure}[!t] 
\centering
\includegraphics[width=0.48\textwidth]{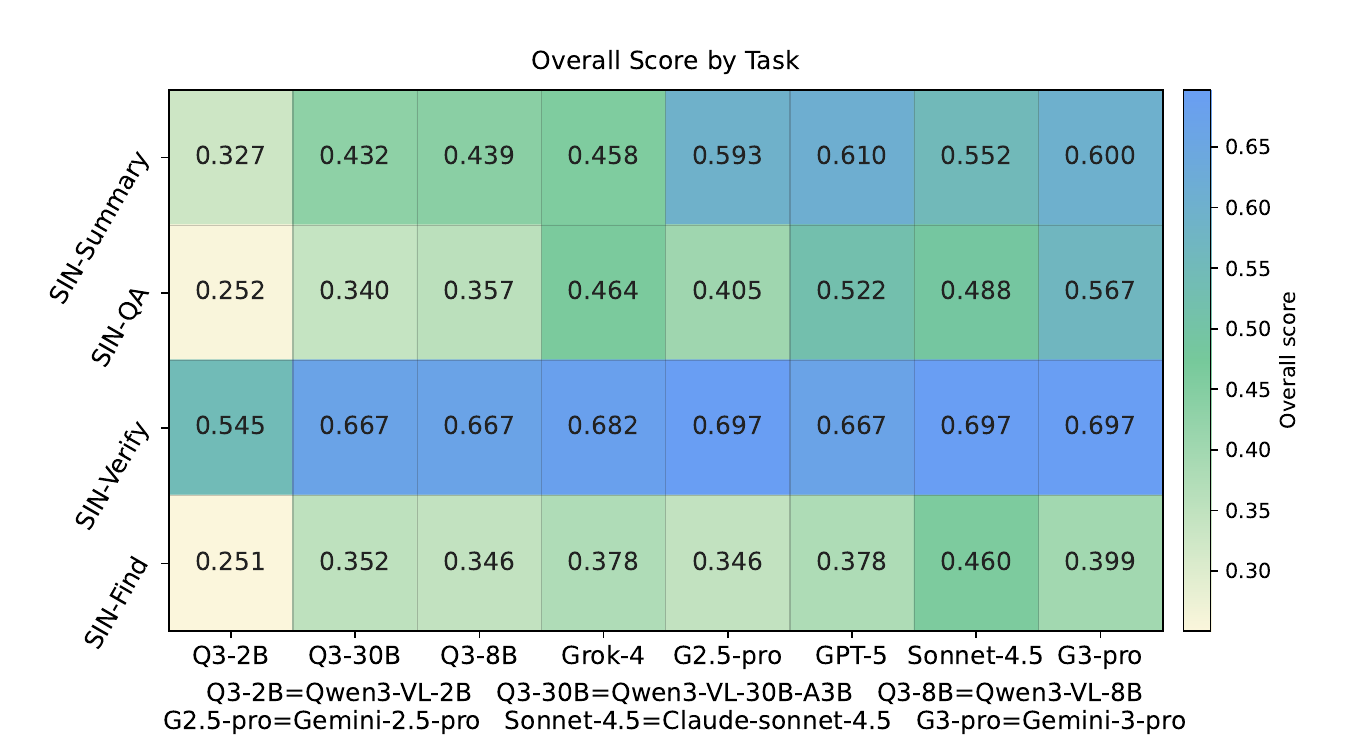} 
\caption{
Task-level overall performance heatmap across models in \texttt{SIN-Bench}. Darker cells imply higher scores.
}
\label{fig:task-result} 
\end{figure}

\subsection{Main Results}
\label{subsec:main_results}

\nbf{Overall Performance.} 
As reported in~\cref{tab:main_results}, Gemini-3-Pro-Preview attains the highest avg. overall score ($0.566$), demonstrating a robust capability for evidence-driven reasoning. 
Although GPT-5 leads in AnsAcc for the SIN-QA task, the overall performance on SIN-Find and SIN-QA (overall: $0.378/0.522$) remains inferior to that of Gemini-3-pro ($0.399/0.567$). 
This discrepancy indicates that Gemini-3-Pro possesses superior capability in grounding reasoning within the multimodal context. 
In contrast, the performance of GPT-5 suggests a reliance on parametric knowledge to infer answers rather than strict adherence to the process of evidence identification.
More analysis are in~\cref{as:add-experiments-analysis}.

Moreover, Qwen3-VL-8B surpasses the larger Qwen3-VL-30B-A3B variant across nearly all metrics. 
This finding suggests that, in the context of scientific reasoning, the density of reasoning-oriented fine-tuning proves more critical than raw parameter count or MoE architectures. 
Additionally, multiple open-source models fail to comply with evidence-formatting constraints. 
This inability leads to invalid scores and underscores the significant challenge associated with generating structured outputs within long multimodal contexts.

\nbf{Task-Level Analysis.} 
As shown in~\cref{fig:task-result}, we report task-level heatmaps across models.

In \textbf{SIN-Find}, Claude-sonnet-4.5 shows superior precision in the identification of scientific anchors (overall: $0.460$), whereas Gemini-2.5-pro excels in logic preservation ($0.536$). 
This pattern indicates a trade-off between precise anchor identification and robust evidence-order preservation.

In \textbf{SIN-Verify}, accuracy remains tightly clustered under the standard setting (Acc=$0.667$--$0.697$ in~\cref{tab:main_results}), but \cref{tab:ver_diag} shows a sharp drop under hard negatives (for example, GPT-5: $1.000 \rightarrow 0.208$; Qwen3-VL-8B: $1.000 \rightarrow 0.044$), which indicates limited logical discrimination for near-miss evidence.
Detailed experiments appear in~\cref{ass:sin-verify-hard-easy}.

\begin{table}[!t]
\centering
\small
\resizebox{0.48\textwidth}{!}{
\begin{tabular}{lccc}
\toprule
\textbf{Setting} & \textbf{Gemini-3-pro} & \textbf{GPT-5} & \textbf{Qwen3-VL-8B} \\
\midrule
Easy Negatives & 1.000 & 1.000 & 1.000 \\
Hard Negatives & \textbf{0.250} & 0.208 & 0.044 \\
\bottomrule
\end{tabular}}
\caption{Accuracy on SIN-Verify under different negative-sampling settings. Hard negatives, which use near-miss evidence, lead to near-chance performance.}
\label{tab:ver_diag}
\end{table}

In \textbf{SIN-QA}, Gemini-3-pro achieves the strongest overall score ($0.567$), whereas GPT-5 delivers the best AnsAcc ($0.767$), which indicates a gap between answer correctness and evidence quality.
This pattern suggests that joint decoding of answers and evidence increases semantic consistency, but it does not guarantee an entailment-support chain.

In \textbf{SIN-Summary}, GPT-5 leads overall score ($0.610$) and achieves the best Logic and Relevance score.
This result suggests that the Logic metric benefits from learning the high-level flow of scientific writing (e.g., Abstract $\rightarrow$ Method $\rightarrow$ Result).

\subsection{Discussion}
\label{ss:discussion}

\begin{figure*}[!t]
\centering
\includegraphics[width=1\textwidth]{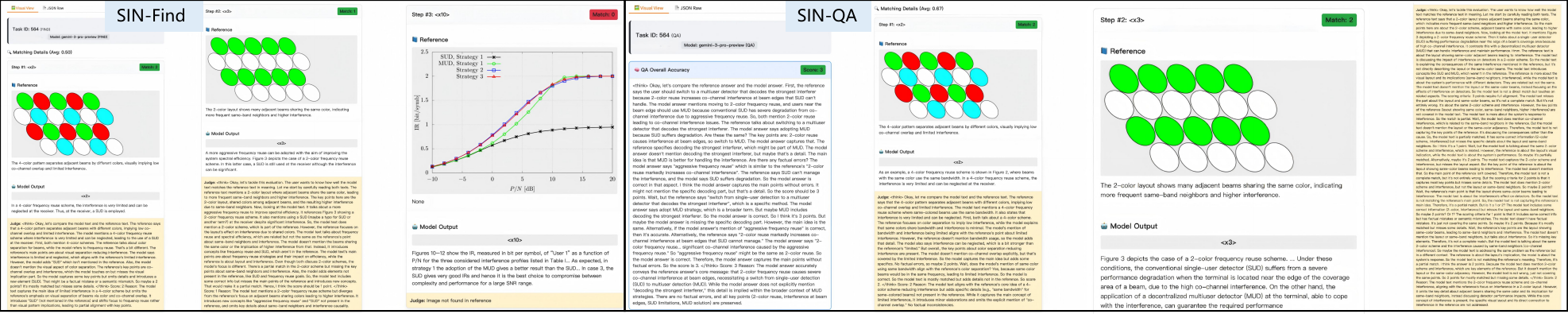}
\caption{Qualitative examples of reasoning failures by Gemini-3-pro in the SIN-Find and SIN-QA tasks.}
\label{fig:error-analysis}
\end{figure*}

\begin{figure}[!t] 
\centering
\includegraphics[width=0.48\textwidth]{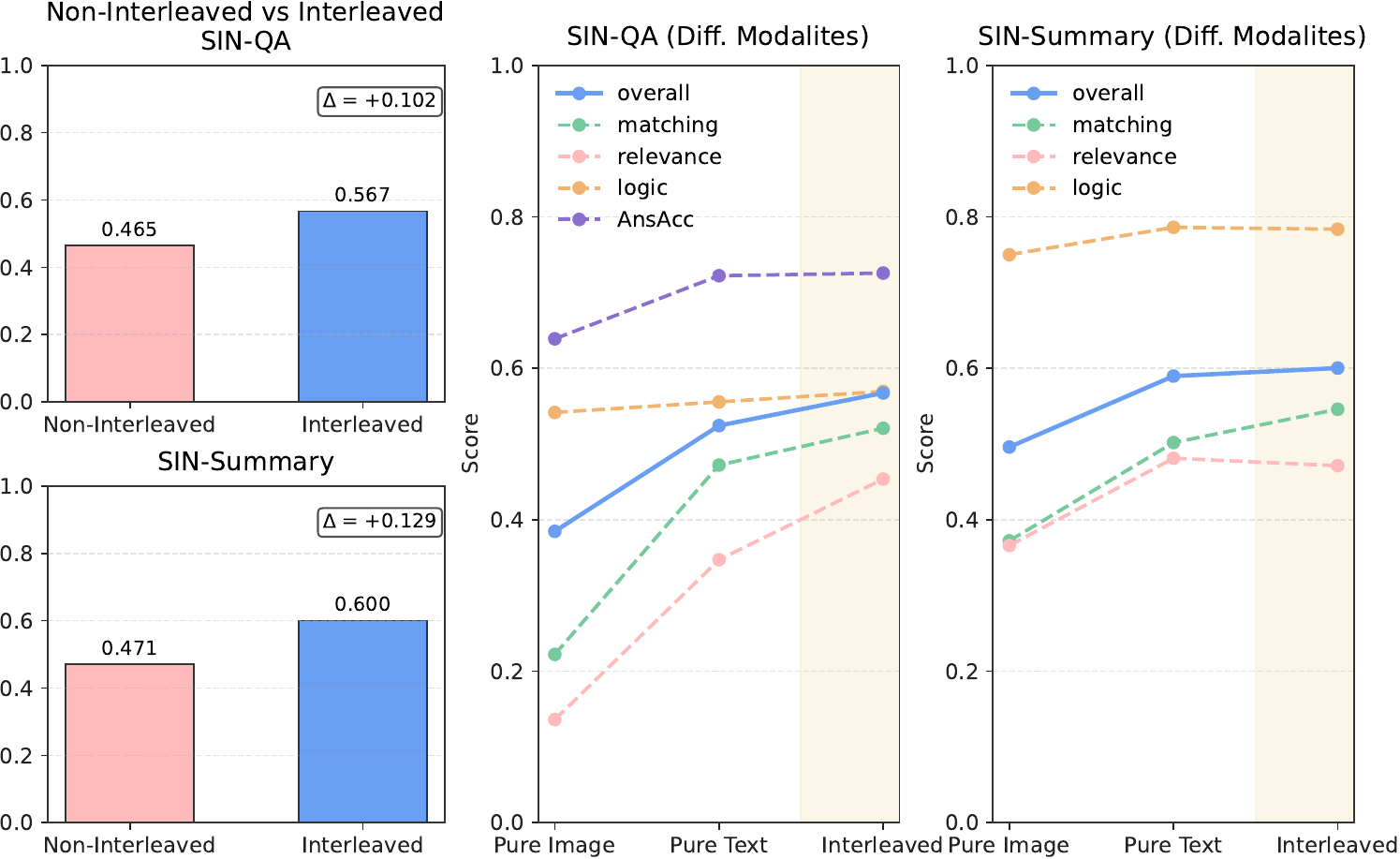} 
\caption{
Impact of interleaved input and modality encodings on SIN-QA and SIN-Summary (Gemini-3-pro).
}
\label{fig:interleaved_analysis} 
\end{figure}

\nbf{Importance of Interleaved Input.}
\cref{fig:interleaved_analysis} shows that preserving the native interleaved structure of scientific papers yields substantial gains over a separated layout (images$\rightarrow$text), improving SIN-QA by $+0.102$ and SIN-Summary by $+0.129$ for Gemini-3-pro.
Moreover, modality ablations indicate ``Interleaved'' $>$ ``Text-only (captions)'' $>$ ``Image-only (rendered pages)'', suggesting that captions retain coarse semantics but lose fine-grained empirical evidence, while raw visuals become most useful when locally grounded by adjacent text.

\nbf{The ``Evidence-chain'' Effect.}
We evaluate whether requiring an explicit evidence chain affects answer correctness by comparing Gemini-3-pro on SIN-QA with and without evidence-chain generation.
In our result, enforcing evidence output improves performance from $0.694$ to $0.726$.
This suggests that evidence-chain generation serves as a lightweight multimodal chain-of-thought, promoting evidence grounding before answering and reducing unsupported guesses.

\begin{figure}[!t] 
\centering
\includegraphics[width=0.48\textwidth]{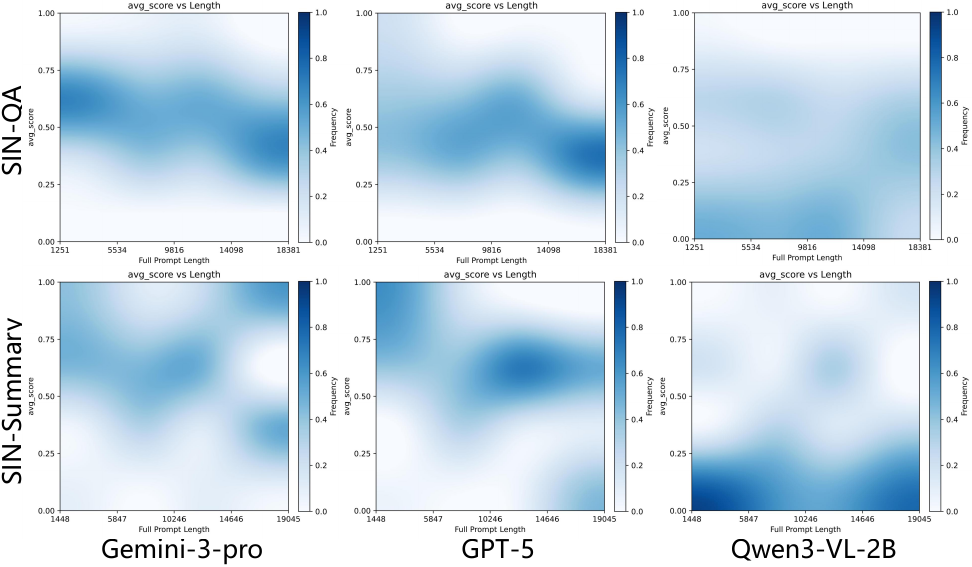} 
\caption{
The score density distribution across varying input token lengths for SIN-QA and SIN-Summary.
}
\label{fig:length_analysis} 
\end{figure}

\nbf{Impact of Text Length on Performance Stability.}
\cref{fig:length_analysis} examines reasoning stability for input contexts exceeding $19$k text tokens (Further analysis in~\cref{ass:impact-doc-len}). 
Both Gemini-3-pro and GPT-5 demonstrate strong adaptability, although GPT-5 exhibits marked tail-end degradation in SIN-QA. 
Conversely, Qwen3-VL-2B displays instability. 
For example, it suffers from a precipitous performance collapse, evidenced by a heavy concentration of scores near zero in SIN-Summary.

\nbf{Error Analysis.}
As shown in~\cref{fig:error-analysis}, our qualitative review identifies two failure modes: 
(1) Information Deficiency, where models omit critical prerequisites in logical chains;
(2) Spurious Reasoning, characterized by irrelevant ``shotgun citations'' that negatively impact precision.
Additionally, we observed that for the same context, when performing SIN-QA, the model tends to provide a more coherent reasoning process and reduces the omission of key information points.
This gap suggests the target answer in SIN-QA acts as a ``semantic anchor,'' guiding the model to track long-range dependencies effectively.
Further analysis is in~\cref{as:case-study}.

\section{Conclusion}

In this work, we introduce the ``Fish-in-the-Ocean'' paradigm and establish \texttt{SIN-Bench}, a comprehensive benchmark tailored for long-context, interleaved scientific reasoning. 
By mandating the construction of explicit evidence chains via the ``No Evidence, No Score'' mechanism, our framework exposes a critical grounding gap in state-of-the-art MLLMs: models like GPT-5 often rely on parametric priors to hallucinate correct answers without genuine contextual comprehension, whereas Gemini-3-pro demonstrates superior adherence to provided evidence.
Furthermore, our analysis identifies significant vulnerabilities in current models, particularly when confronting adversarial ``hard negatives'' or adhering to structured output constraints in open-weight architectures.
Ultimately, this study advocates for a fundamental shift in multimodal evaluation—moving beyond mere response accuracy to rigorously assess the traceability and logic of the reasoning process. 

\section*{Limitations}

The breadth of the evaluation faces constraints due to the input requirements of current multimodal architectures. 
While the study incorporates leading generalist models such as GPT-5 and Gemini-3-Pro, the scarcity of models supporting long-context interleaved inputs prevents the inclusion of experts in specific domains.
To mitigate this, we \textbf{open-source all data construction and evaluation codes}. 
This standardized interface assists the community in the integration of emerging MLLMs, thereby expanding future evaluation coverage.

Furthermore, strict filtering strategies designed to minimize parsing noise result in a trade-off regarding data utilization, as documents with minor imperfections are inadvertently excluded. 
Fortunately, the pipeline features highly configurable filtering rules. 
This flexibility allows users to adjust thresholds and voting mechanisms, enabling an optimal balance between purity and scale to retrieve valuable documents.

\section*{Ethical considerations}

We adhere strictly to data compliance principles. 
All documents in \texttt{SIN-Data} originate from open-access repositories, specifically arXiv and PMC, and we comply with relevant licensing agreements such as CC-BY 4.0.
However, the risk of misuse warrants attention, as entities such as ``paper mills''~\citep{DBLP:journals/lp/KendallS24@risk-llm-chatgpt} may exploit this technology to generate fabricated papers. 
While the ``No Evidence, No Score'' principle serves as a verification mechanism, we acknowledge the potential for reverse exploitation to create realistic fabrications with plausible evidence chains. 
Consequently, we urge the community to shift the research focus from generation to the utilization of evidence chains for the detection of academic fraud, and we encourage the disclosure of technical details to promote transparency.

\bibliography{custom}

@article{DBLP:journals/corr/abs-2406-13923@pin,
  author       = {Junjie Wang and
                  Yin Zhang and
                  Yatai Ji and
                  Yuxiang Zhang and
                  Chunyang Jiang and
                  Yubo Wang and
                  Kang Zhu and
                  Zekun Wang and
                  Tiezhen Wang and
                  Wenhao Huang and
                  Jie Fu and
                  Bei Chen and
                  Qunshu Lin and
                  Minghao Liu and
                  Ge Zhang and
                  Wenhu Chen},
  title        = {{PIN:} {A} Knowledge-Intensive Dataset for Paired and Interleaved
                  Multimodal Documents},
  journal      = {CoRR},
  volume       = {abs/2406.13923},
  year         = {2024}
}

@inproceedings{DBLP:conf/nips/WangZRDLLH0ZLZL24@mm-niah,
  author       = {Weiyun Wang and
                  Shuibo Zhang and
                  Yiming Ren and
                  Yuchen Duan and
                  Tiantong Li and
                  Shuo Liu and
                  Mengkang Hu and
                  Zhe Chen and
                  Kaipeng Zhang and
                  Lewei Lu and
                  Xizhou Zhu and
                  Ping Luo and
                  Yu Qiao and
                  Jifeng Dai and
                  Wenqi Shao and
                  Wenhai Wang},
  title        = {Needle In {A} Multimodal Haystack},
  booktitle    = {NeurIPS},
  year         = {2024}
}

@article{DBLP:journals/corr/abs-2406-11633@DocGenome,
  author       = {Renqiu Xia and
                  Song Mao and
                  Xiangchao Yan and
                  Hongbin Zhou and
                  Bo Zhang and
                  Haoyang Peng and
                  Jiahao Pi and
                  Daocheng Fu and
                  Wenjie Wu and
                  Hancheng Ye and
                  Shiyang Feng and
                  Bin Wang and
                  Chao Xu and
                  Conghui He and
                  Pinlong Cai and
                  Min Dou and
                  Botian Shi and
                  Sheng Zhou and
                  Yongwei Wang and
                  Bin Wang and
                  Junchi Yan and
                  Fei Wu and
                  Yu Qiao},
  title        = {DocGenome: An Open Large-scale Scientific Document Benchmark for Training
                  and Testing Multi-modal Large Language Models},
  journal      = {CoRR},
  volume       = {abs/2406.11633},
  year         = {2024}
}

@inproceedings{DBLP:conf/nips/MaZC0JLLLMDZP0W24@mmlongbench-doc,
  author       = {Yubo Ma and
                  Yuhang Zang and
                  Liangyu Chen and
                  Meiqi Chen and
                  Yizhu Jiao and
                  Xinze Li and
                  Xinyuan Lu and
                  Ziyu Liu and
                  Yan Ma and
                  Xiaoyi Dong and
                  Pan Zhang and
                  Liangming Pan and
                  Yu{-}Gang Jiang and
                  Jiaqi Wang and
                  Yixin Cao and
                  Aixin Sun},
  title        = {{MMLONGBENCH-DOC:} Benchmarking Long-context Document Understanding
                  with Visualizations},
  booktitle    = {NeurIPS},
  year         = {2024}
}

@inproceedings{DBLP:conf/acl/DengYBWLXLGS0025@LongDocURL,
  author       = {Chao Deng and
                  Jiale Yuan and
                  Pi Bu and
                  Peijie Wang and
                  Zhong{-}Zhi Li and
                  Jian Xu and
                  Xiao{-}Hui Li and
                  Yuan Gao and
                  Jun Song and
                  Bo Zheng and
                  Cheng{-}Lin Liu},
  title        = {LongDocURL: a Comprehensive Multimodal Long Document Benchmark Integrating
                  Understanding, Reasoning, and Locating},
  booktitle    = {{ACL} {(1)}},
  pages        = {1135--1159},
  publisher    = {Association for Computational Linguistics},
  year         = {2025}
}

@article{@Qwen3-VL,
      title={Qwen3-VL Technical Report}, 
      author={Shuai Bai and Yuxuan Cai and Ruizhe Chen and Keqin Chen and Xionghui Chen and Zesen Cheng and Lianghao Deng and Wei Ding and Chang Gao and Chunjiang Ge and Wenbin Ge and Zhifang Guo and Qidong Huang and Jie Huang and Fei Huang and Binyuan Hui and Shutong Jiang and Zhaohai Li and Mingsheng Li and Mei Li and Kaixin Li and Zicheng Lin and Junyang Lin and Xuejing Liu and Jiawei Liu and Chenglong Liu and Yang Liu and Dayiheng Liu and Shixuan Liu and Dunjie Lu and Ruilin Luo and Chenxu Lv and Rui Men and Lingchen Meng and Xuancheng Ren and Xingzhang Ren and Sibo Song and Yuchong Sun and Jun Tang and Jianhong Tu and Jianqiang Wan and Peng Wang and Pengfei Wang and Qiuyue Wang and Yuxuan Wang and Tianbao Xie and Yiheng Xu and Haiyang Xu and Jin Xu and Zhibo Yang and Mingkun Yang and Jianxin Yang and An Yang and Bowen Yu and Fei Zhang and Hang Zhang and Xi Zhang and Bo Zheng and Humen Zhong and Jingren Zhou and Fan Zhou and Jing Zhou and Yuanzhi Zhu and Ke Zhu},
      journal={arXiv preprint arXiv:2511.21631},
      year={2025}
}

@techreport{openai2025@gpt-5,
  title       = {GPT-5 System Card},
  author      = {OpenAI},
  year        = {2025},
  institution = {OpenAI},
  url         = {https://cdn.openai.com/gpt-5-system-card.pdf}
}

@techreport{google2025@gemini-3-pro,
  title       = {Gemini 3 Pro Model Card},
  author      = {Google DeepMind},
  year        = {2025},
  institution = {Google},
  url         = {https://storage.googleapis.com/deepmind-media/Model-Cards/Gemini-3-Pro-Model-Card.pdf}
}

@techreport{xai2025@grok-4,
  title       = {Grok-4 Model Card},
  author      = {xAI},
  institution = {xAI},
  year        = {2025},
  month       = {August},
  type        = {Model Card},
  url         = {https://data.x.ai/2025-08-20-grok-4-model-card.pdf}
}

@inproceedings{DBLP:conf/wacv/MathewKJ21@DocVQA,
  author       = {Minesh Mathew and
                  Dimosthenis Karatzas and
                  C. V. Jawahar},
  title        = {DocVQA: {A} Dataset for {VQA} on Document Images},
  booktitle    = {{WACV}},
  pages        = {2199--2208},
  publisher    = {{IEEE}},
  year         = {2021}
}

@inproceedings{DBLP:conf/naacl/DasigiLBCSG21@qasper,
  author       = {Pradeep Dasigi and
                  Kyle Lo and
                  Iz Beltagy and
                  Arman Cohan and
                  Noah A. Smith and
                  Matt Gardner},
  title        = {A Dataset of Information-Seeking Questions and Answers Anchored in
                  Research Papers},
  booktitle    = {{NAACL-HLT}},
  pages        = {4599--4610},
  publisher    = {Association for Computational Linguistics},
  year         = {2021}
}

@article{DBLP:journals/corr/abs-2510-13276@MMLongCite,
  author       = {Keyan Zhou and
                  Zecheng Tang and
                  Lingfeng Ming and
                  Guanghao Zhou and
                  Qiguang Chen and
                  Dan Qiao and
                  Zheming Yang and
                  Libo Qin and
                  Minghui Qiu and
                  Juntao Li and
                  Min Zhang},
  title        = {MMLongCite: {A} Benchmark for Evaluating Fidelity of Long-Context
                  Vision-Language Models},
  journal      = {CoRR},
  volume       = {abs/2510.13276},
  year         = {2025}
}

@inproceedings{DBLP:conf/icdar/ZhongTJ19@PubLayNet,
  author       = {Xu Zhong and
                  Jianbin Tang and
                  Antonio Jimeno{-}Yepes},
  title        = {PubLayNet: Largest Dataset Ever for Document Layout Analysis},
  booktitle    = {{ICDAR}},
  pages        = {1015--1022},
  publisher    = {{IEEE}},
  year         = {2019}
}

@inproceedings{DBLP:conf/iclr/XiaHQZWZCCDLWY25@MMIE,
  author       = {Peng Xia and
                  Siwei Han and
                  Shi Qiu and
                  Yiyang Zhou and
                  Zhaoyang Wang and
                  Wenhao Zheng and
                  Zhaorun Chen and
                  Chenhang Cui and
                  Mingyu Ding and
                  Linjie Li and
                  Lijuan Wang and
                  Huaxiu Yao},
  title        = {{MMIE:} Massive Multimodal Interleaved Comprehension Benchmark for
                  Large Vision-Language Models},
  booktitle    = {{ICLR}},
  publisher    = {OpenReview.net},
  year         = {2025}
}

@article{DBLP:journals/corr/abs-2503-21839@M-DocSum,
  author       = {Haolong Yan and
                  Kaijun Tan and
                  Yeqing Shen and
                  Xin Huang and
                  Zheng Ge and
                  Xiangyu Zhang and
                  Si Li and
                  Daxin Jiang},
  title        = {M-DocSum: Do LVLMs Genuinely Comprehend Interleaved Image-Text in
                  Document Summarization?},
  journal      = {CoRR},
  volume       = {abs/2503.21839},
  year         = {2025}
}

@article{DBLP:journals/corr/abs-2409-12959@MMSearch,
  author       = {Dongzhi Jiang and
                  Renrui Zhang and
                  Ziyu Guo and
                  Yanmin Wu and
                  Jiayi Lei and
                  Pengshuo Qiu and
                  Pan Lu and
                  Zehui Chen and
                  Guanglu Song and
                  Peng Gao and
                  Yu Liu and
                  Chunyuan Li and
                  Hongsheng Li},
  title        = {MMSearch: Benchmarking the Potential of Large Models as Multi-modal
                  Search Engines},
  journal      = {CoRR},
  volume       = {abs/2409.12959},
  year         = {2024}
}

@article{DBLP:journals/corr/abs-2506-02454@DeepResearcher,
  author       = {Zhaorui Yang and
                  Bo Pan and
                  Han Wang and
                  Yiyao Wang and
                  Xingyu Liu and
                  Minfeng Zhu and
                  Bo Zhang and
                  Wei Chen},
  title        = {Multimodal DeepResearcher: Generating Text-Chart Interleaved Reports
                  From Scratch with Agentic Framework},
  journal      = {CoRR},
  volume       = {abs/2506.02454},
  year         = {2025}
}

@inproceedings{DBLP:conf/iclr/BlecherCSS24@nougat,
  author       = {Lukas Blecher and
                  Guillem Cucurull and
                  Thomas Scialom and
                  Robert Stojnic},
  title        = {Nougat: Neural Optical Understanding for Academic Documents},
  booktitle    = {{ICLR}},
  publisher    = {OpenReview.net},
  year         = {2024}
}

@misc{meta2025llama4multimodal@llama-4,
  author       = {{AI at Meta}},
  title        = {{The Llama 4 herd: The beginning of a new era of natively multimodal AI innovation}},
  year         = {2025},
  month        = apr,
  howpublished = {\url{https://ai.meta.com/blog/llama-4-multimodal-intelligence/}},
  note         = {Published April 5, 2025. Accessed 2025-12-30}
}

@inproceedings{DBLP:conf/nips/LaurenconSTBSLW23@OBELICS,
  author       = {Hugo Lauren{\c{c}}on and
                  Lucile Saulnier and
                  L{\'{e}}o Tronchon and
                  Stas Bekman and
                  Amanpreet Singh and
                  Anton Lozhkov and
                  Thomas Wang and
                  Siddharth Karamcheti and
                  Alexander M. Rush and
                  Douwe Kiela and
                  Matthieu Cord and
                  Victor Sanh},
  title        = {{OBELICS:} An Open Web-Scale Filtered Dataset of Interleaved Image-Text
                  Documents},
  booktitle    = {NeurIPS},
  year         = {2023}
}

@article{DBLP:journals/corr/abs-2508-15763@Intern-S1,
  author       = {Lei Bai and
                  Zhongrui Cai and
                  Yuhang Cao and
                  Maosong Cao and
                  Weihan Cao and
                  Chiyu Chen and
                  Haojiong Chen and
                  Kai Chen and
                  Pengcheng Chen and
                  Ying Chen and
                  Yongkang Chen and
                  Yu Cheng and
                  Pei Chu and
                  Tao Chu and
                  Erfei Cui and
                  Ganqu Cui and
                  Long Cui and
                  Ziyun Cui and
                  Nianchen Deng and
                  Ning Ding and
                  Nanqing Dong and
                  Peijie Dong and
                  Shihan Dou and
                  Sinan Du and
                  Haodong Duan and
                  Caihua Fan and
                  Ben Gao and
                  Changjiang Gao and
                  Jianfei Gao and
                  Songyang Gao and
                  Yang Gao and
                  Zhangwei Gao and
                  Jiaye Ge and
                  Qiming Ge and
                  Lixin Gu and
                  Yuzhe Gu and
                  Aijia Guo and
                  Qipeng Guo and
                  Xu Guo and
                  Conghui He and
                  Junjun He and
                  Yili Hong and
                  Siyuan Hou and
                  Caiyu Hu and
                  Hanglei Hu and
                  Jucheng Hu and
                  Ming Hu and
                  Zhouqi Hua and
                  Haian Huang and
                  Junhao Huang and
                  Xu Huang and
                  Zixian Huang and
                  Zhe Jiang and
                  Lingkai Kong and
                  Linyang Li and
                  Peiji Li and
                  Pengze Li and
                  Shuaibin Li and
                  Tianbin Li and
                  Wei Li and
                  Yuqiang Li and
                  Dahua Lin and
                  Junyao Lin and
                  Tianyi Lin and
                  Zhishan Lin and
                  Hongwei Liu and
                  Jiangning Liu and
                  Jiyao Liu and
                  Junnan Liu and
                  Kai Liu and
                  Kaiwen Liu and
                  Kuikun Liu and
                  Shichun Liu and
                  Shudong Liu and
                  Wei Liu and
                  Xinyao Liu and
                  Yuhong Liu and
                  Zhan Liu and
                  Yinquan Lu and
                  Haijun Lv and
                  Hongxia Lv and
                  Huijie Lv and
                  Qitan Lv and
                  Ying Lv and
                  Chengqi Lyu and
                  Chenglong Ma and
                  Jianpeng Ma and
                  Ren Ma and
                  Runmin Ma and
                  Runyuan Ma and
                  Xinzhu Ma and
                  Yichuan Ma and
                  Zihan Ma and
                  Sixuan Mi and
                  Junzhi Ning and
                  Wenchang Ning and
                  Xinle Pang and
                  Jiahui Peng and
                  Runyu Peng and
                  Yu Qiao},
  title        = {Intern-S1: {A} Scientific Multimodal Foundation Model},
  journal      = {CoRR},
  volume       = {abs/2508.15763},
  year         = {2025}
}

@article{DBLP:journals/corr/abs-2507-06261@gemini-2.5,
  author       = {Gemini Team},
  title        = {Gemini 2.5: Pushing the Frontier with Advanced Reasoning, Multimodality,
                  Long Context, and Next Generation Agentic Capabilities},
  journal      = {CoRR},
  volume       = {abs/2507.06261},
  year         = {2025}
}

@techreport{anthropic2025@claude-4.5,
  title = {Claude Sonnet 4.5 System Card},
  author = {Anthropic},
  year = {2025},
  month = {October},
  institution = {Anthropic},
  url = {https://www.anthropic.com/claude-sonnet-4-5-system-card},
  type = {System Card}
}

@article{kendall1948rank@kendall,
  title={Rank correlation methods.},
  author={Kendall, Maurice George},
  year={1948},
  publisher={Griffin}
}

@article{DBLP:journals/lp/KendallS24@risk-llm-chatgpt,
  author       = {Graham Kendall and
                  Jaime A. Teixeira da Silva},
  title        = {Risks of abuse of large language models, like ChatGPT, in scientific
                  publishing: Authorship, predatory publishing, and paper mills},
  journal      = {Learn. Publ.},
  volume       = {37},
  number       = {1},
  pages        = {55--62},
  year         = {2024}
}

@inproceedings{DBLP:conf/emnlp/LiuIXWXZ23@g-eval,
  author       = {Yang Liu and
                  Dan Iter and
                  Yichong Xu and
                  Shuohang Wang and
                  Ruochen Xu and
                  Chenguang Zhu},
  title        = {G-Eval: {NLG} Evaluation using Gpt-4 with Better Human Alignment},
  booktitle    = {{EMNLP}},
  pages        = {2511--2522},
  publisher    = {Association for Computational Linguistics},
  year         = {2023}
}

@inproceedings{DBLP:conf/nips/ZhengC00WZL0LXZ23@mt-bench,
  author       = {Lianmin Zheng and
                  Wei{-}Lin Chiang and
                  Ying Sheng and
                  Siyuan Zhuang and
                  Zhanghao Wu and
                  Yonghao Zhuang and
                  Zi Lin and
                  Zhuohan Li and
                  Dacheng Li and
                  Eric P. Xing and
                  Hao Zhang and
                  Joseph E. Gonzalez and
                  Ion Stoica},
  title        = {Judging LLM-as-a-Judge with MT-Bench and Chatbot Arena},
  booktitle    = {NeurIPS},
  year         = {2023}
}

@article{DBLP:journals/corr/abs-2507-15882@document-haystack,
  author       = {Goeric Huybrechts and
                  Srikanth Ronanki and
                  Sai Muralidhar Jayanthi and
                  Jack FitzGerald and
                  Srinivasan Veeravanallur},
  title        = {Document Haystack: {A} Long Context Multimodal Image/Document Understanding
                  Vision {LLM} Benchmark},
  journal      = {CoRR},
  volume       = {abs/2507.15882},
  year         = {2025}
}

@inproceedings{DBLP:conf/nips/00040HA24@SciFIBench,
  author       = {Jonathan Roberts and
                  Kai Han and
                  Neil Houlsby and
                  Samuel Albanie},
  title        = {SciFIBench: Benchmarking Large Multimodal Models for Scientific Figure
                  Interpretation},
  booktitle    = {NeurIPS},
  year         = {2024}
}

@article{DBLP:journals/corr/abs-2505-09388@qwen3,
  author       = {An Yang and
                  Anfeng Li and
                  Baosong Yang and
                  Beichen Zhang and
                  Binyuan Hui and
                  Bo Zheng and
                  Bowen Yu and
                  Chang Gao and
                  Chengen Huang and
                  Chenxu Lv and
                  Chujie Zheng and
                  Dayiheng Liu and
                  Fan Zhou and
                  Fei Huang and
                  Feng Hu and
                  Hao Ge and
                  Haoran Wei and
                  Huan Lin and
                  Jialong Tang and
                  Jian Yang and
                  Jianhong Tu and
                  Jianwei Zhang and
                  Jian Yang and
                  Jiaxi Yang and
                  Jingren Zhou and
                  Junyang Lin and
                  Kai Dang and
                  Keqin Bao and
                  Kexin Yang and
                  Le Yu and
                  Lianghao Deng and
                  Mei Li and
                  Mingfeng Xue and
                  Mingze Li and
                  Pei Zhang and
                  Peng Wang and
                  Qin Zhu and
                  Rui Men and
                  Ruize Gao and
                  Shixuan Liu and
                  Shuang Luo and
                  Tianhao Li and
                  Tianyi Tang and
                  Wenbiao Yin and
                  Xingzhang Ren and
                  Xinyu Wang and
                  Xinyu Zhang and
                  Xuancheng Ren and
                  Yang Fan and
                  Yang Su and
                  Yichang Zhang and
                  Yinger Zhang and
                  Yu Wan and
                  Yuqiong Liu and
                  Zekun Wang and
                  Zeyu Cui and
                  Zhenru Zhang and
                  Zhipeng Zhou and
                  Zihan Qiu},
  title        = {Qwen3 Technical Report},
  journal      = {CoRR},
  volume       = {abs/2505.09388},
  year         = {2025}
}

@article{DBLP:journals/chinaf/ChenWTYGCTHLMMWDYGHSJXW24@how-far,
  author       = {Zhe Chen and
                  Weiyun Wang and
                  Hao Tian and
                  Shenglong Ye and
                  Zhangwei Gao and
                  Erfei Cui and
                  Wenwen Tong and
                  Kongzhi Hu and
                  Jiapeng Luo and
                  Zheng Ma and
                  Ji Ma and
                  Jiaqi Wang and
                  Xiaoyi Dong and
                  Hang Yan and
                  Hewei Guo and
                  Conghui He and
                  Botian Shi and
                  Zhenjiang Jin and
                  Chao Xu and
                  Bin Wang and
                  Xingjian Wei and
                  Wei Li and
                  Wenjian Zhang and
                  Bo Zhang and
                  Pinlong Cai and
                  Licheng Wen and
                  Xiangchao Yan and
                  Min Dou and
                  Lewei Lu and
                  Xizhou Zhu and
                  Tong Lu and
                  Dahua Lin and
                  Yu Qiao and
                  Jifeng Dai and
                  Wenhai Wang},
  title        = {How far are we to GPT-4V? Closing the gap to commercial multimodal
                  models with open-source suites},
  journal      = {Sci. China Inf. Sci.},
  volume       = {67},
  number       = {12},
  year         = {2024},
  url          = {https://doi.org/10.1007/s11432-024-4231-5},
  doi          = {10.1007/S11432-024-4231-5},
  bibsource    = {dblp computer science bibliography, https://dblp.org}
}

@article{DBLP:journals/corr/abs-2411-15296@mme-survey,
  author       = {Chaoyou Fu and
                  Yifan Zhang and
                  Shukang Yin and
                  Bo Li and
                  Xinyu Fang and
                  Sirui Zhao and
                  Haodong Duan and
                  Xing Sun and
                  Ziwei Liu and
                  Liang Wang and
                  Caifeng Shan and
                  Ran He},
  title        = {MME-Survey: {A} Comprehensive Survey on Evaluation of Multimodal LLMs},
  journal      = {CoRR},
  volume       = {abs/2411.15296},
  year         = {2024}
}

\clearpage
\appendix
\section*{Appendix}

\section{Details of Benchmark Creation}

\subsection{SIN-Data Construction Pipeline}
\label{ass:sin-data-pipeline}

\nbf{Design Philosophy.}
We strategically target the Scientific INterleaved (SIN) format to address the scarcity of high-fidelity reasoning data in current multimodal training and evaluation. 
While large-scale datasets like OBELICS~\citep{DBLP:conf/nips/LaurenconSTBSLW23@OBELICS} provide vast interleaved sequences, they primarily originate from general web sources where the semantic alignment between text and images is often loose or incidental.
In contrast, scientific literature exhibits a rigorous symbiotic dependency where the text provides analytical narratives and figures offer empirical evidence. 
This domain offers a unique advantage in content density because scientific writing is inherently knowledge-intensive with a high signal-to-noise ratio. 
It demands precise terminology and logic rather than generic descriptions.

Furthermore, the interleaved structure is critical for fostering long-context reasoning capabilities. 
In the scientific context, the interpretation of a single figure often necessitates synthesizing information scattered across pages of dense prose. 
We preserve this non-linear ``chain of evidence'' by logically embedding visuals within the narrative. 
This design compels models to perform multi-hop reasoning across extended contexts, a capability often absent in standard caption-centric datasets.

To harness this potential, the SIN-Data pipeline is engineered to bridge the structural heterogeneity between disciplines, specifically between the LaTeX-based source packages predominant in Physics and Computer Science and the XML-structured archives typical of Biomedicine. 
Our pipeline unifies these disparate sources into a standardized format while offering granular controllability. 
By computing rich quality signals during the homogenization process, such as token length and visual density, we transform the dataset from a raw collection into a structured resource. 
This enables precise stratification and analysis of model performance across varying degrees of multimodal complexity.

We draw inspiration from the PIN dataset~\citep{DBLP:journals/corr/abs-2406-13923@pin} to re-engineer the parsing workflow and include the quality signals for constructing our \texttt{SIN-Data} dataset.
As shown in~\cref{fig:sin-data-bench} (a), our workflow operates in three progressive stages:

\subsubsection{Stage 1: Element Parsing} 
The initial phase focuses on decoupling content from presentation to generate intermediate structured files.
\begin{itemize}[nosep, leftmargin=*]
\item arXiv Stream: We compile LaTeX source packages into responsive HTML using \texttt{Engrafo}\footnote{\url{https://github.com/arxiv-vanity/engrafo}}, followed by NOUGAT~\citep{DBLP:conf/iclr/BlecherCSS24@nougat} parsing to extract text. A visual matching algorithm recovers image references from the DOM tree, re-anchoring them to their textual contexts.
\item PMC Stream: We leverage a customized \texttt{s2orc-doc2json}\footnote{\url{https://github.com/allenai/s2orc-doc2json}} parser to process JATS XML files. This module robustly extracts core academic elements—discarding stylistic markup while preserving reference links and tables—to output structured JSON files.
\end{itemize}

\subsubsection{Stage 2: Semantic-First Formatting \& Signal Extraction}
We transform the intermediate files into a Unified Interleaved Markdown format. A core innovation is our citation-driven injection strategy, which inserts a unique image placeholder $\langle x_k \rangle$ immediately preceding the paragraph where the image is first cited, preserving the logical ``chain of evidence'' rather than spatial layout.
Moreover, during this linearization process, we compute a suite of granular quality signals for each document. 
\begin{itemize}[nosep, leftmargin=*]
\item Volume \& Granularity: We record total\_tokens (calculated via the Llama-4 tokenizer~\citep{meta2025llama4multimodal@llama-4}) and avg\_segment\_length (the average number of tokens per text block) to measure context length and textual density.
\item Visual Density: We calculate image\_count and image\_ratio to quantify the proportion of visual information.
\item Interconnectivity: We track interleave\_segments, representing the frequency of text-image intersections, which serves as a proxy for the complexity of the multimodal reasoning chain.
\item Layout \& Resolution: We also extract format-specific features, such as column layout (single vs. double) and average image resolution, to assess visual quality. These signals are embedded as metadata, serving as the quantitative basis for the subsequent filtering.
\end{itemize}

\subsubsection{Stage 3: Quality Filtering \& Taxonomy Alignment} 
By utilizing the quality signals computed in Stage 2, we curate the dataset based on the principle of High-Density Multimodal Interconnectivity. 
We filter out samples with sparse visual contexts, broken reference chains, or extreme lengths (retaining $32$k–$1$M tokens).
To ensure broad domain coverage, we reference the arXiv taxonomy to categorize and annotate all samples with Qwen3-VL-2B~\citep{@Qwen3-VL} (Details are in~\cref{ass:discipline-taxonomy}). 
We employ a stratified sampling strategy across these categories to prevent domain collapse, ensuring the final dataset retains a representative distribution across diverse scientific fields.

From an initial collection of $50,000$ source packages, our rigorous filtering pipeline yielded a curated set of $4,000$ high-quality documents. 
This final corpus spans $10+$ top-level disciplines and over $80+$ subfields, providing a diverse and structurally controllable testbed for scientific reasoning.

\subsection{SIN-Bench Construction Pipeline}
\label{ass:sin-bench-pipeline}

Building upon the interleaved document representations from \texttt{SIN-Data}, we establish an iterative human--model collaborative pipeline for the construction of \texttt{SIN-Bench}.
This framework adheres to the principle of minimizing human intervention while strictly maintaining the validity of samples.
As shown in~\cref{fig:sin-data-bench} (c), to achieve scalability and reliability, we employ structured synthesis alongside cross-model consistency checks and Human-in-the-Loop protocols.
These measures ensure factual correctness, logical self-consistency, and the verifiability of evidence.
As illustrated, the workflow comprises four sequential stages and a self-improving loop: Seed Example Collection $\rightarrow$ (1) Multi-Model Synthesis $\rightarrow$ (2) Cross-Verification $\rightarrow$ (3) Human-in-the-Loop Audit $\rightarrow$ Golden-to-Seed Iteration.

\nbf{Input Preparation: Pruning and Alignment.}
Given the extensive length typical of scientific literature, we perform strategic pruning to generate input segments compatible with the context limits of models, while strictly preserving the logical coupling of the SIN format.
We prioritize the retention of core sections—such as the Abstract, Introduction, Methods, Experiments, and Conclusion—and preserve text spans adjacent to the first citation of figures or tables, alongside their corresponding visual anchors $\langle x_k\rangle$.
This process guarantees that key claims and their associated evidence anchors remain locatable and verifiable, even under the compression of the input.

\nbf{Seed Examples.}
We manually curate a small set of high-quality seed examples for each of the four tasks to standardize the output format and the interface of the evidence chain.

\nbf{(1) Multi-MLLM Collaborative Synthesis.}
Given a single document $D$, we leverage a diverse array of MLLMs to collaboratively synthesize candidate samples, which includes Gemini-3-pro, GPT-5, Qwen3-VL, Grok-4, and Claude-4.5.
We designate SIN-QA and SIN-Summary as the core generation pivot.
By requiring models to co-generate the content (answers or summaries) and the evidence chain $E$ within a unified context window, we mitigate the semantic drift and lack of self-consistency often inherent in traditional pipelines.
Subsequently, we perform synchronous task derivation within the \emph{same synthesis session} to structurally guarantee cross-task consistency:
\begin{itemize}
\item \textbf{SIN-QA}: The model cgenerates the answer $A$ and the evidence chain $E$ for a specific reasoning query $Q$.
This process ensures strict semantic alignment between the generated rationale and the source substrate.
\item \textbf{SIN-Find}: We reformulate the valid outputs from the core tasks into evidence localization challenges. 
By fixing the query $Q$ or a specific claim from the summary, we require the reconstruction of the supporting evidence chain.
\item \textbf{SIN-Verify}: We construct discriminative instances by applying negative sampling to the valid chains. 
We introduce ``Insufficient Evidence'' (pairing valid claims with unrelated evidence) and ``Perturbed Evidence'' (disrupting the order or alignment of valid chains) to test the capability of the model to audit logical gaps.
\item \textbf{SIN-Summary}: Adopting a ``cite-as-you-write'' strategy, the model produces a holistic summary where each key statement explicitly links to verifiable evidence anchors in $D$, thereby operationalizing grounded long-context synthesis.
\end{itemize}

This phase produces a pool of candidate samples covering all four tasks. 
It maintains structural coupling across samples, which facilitates unified quality control in subsequent steps.


\begin{figure*}[!tb]
\small
\begin{tcolorbox}[
    colback=promptbg,
    colframe=promptframe,
    boxrule=0.5mm,
    arc=0mm,
    outer arc=0mm,
    title=Cross-Validation Auditor Prompt: SIN-Summary
]
Role:\\
You are a rigorous academic reviewer auditing the \textbf{evidence quality} of a summary generated from a research paper. Your judgment must be \textbf{evidence-centric} and \textbf{strictly grounded} in the provided excerpt.\\

Inputs (you will receive):\\
1) A Markdown excerpt from a research paper (may contain \texttt{<figure>} placeholders corresponding to images/figures).\\
2) A list of summary evidence items. Each item contains (i) an image anchor marker (e.g., \texttt{x1}, \texttt{x2}, \dots) and (ii) a text snippet describing key points.\\

Task Description:\\
Evaluate whether the provided evidence items faithfully support the claimed summary points, with explicit verification of \textbf{anchor-to-content alignment} (i.e., whether each \texttt{xk} correctly maps to relevant visual content and its surrounding text context in the excerpt).\\

Rubric (Criterionized, 1--5):\\
\textbf{Evidence Quality} (integer score only)\\
- \textbf{5}: Faithful and complete support; anchors are correctly matched; coverage includes key aspects (method, results, conclusions); organization is coherent.\\
- \textbf{4}: Mostly faithful with minor omissions or slightly weak anchor justification; no critical mismatches.\\
- \textbf{3}: Partial support; noticeable gaps in coverage or unclear anchor relevance; some evidence is loosely connected.\\
- \textbf{2}: Weak support; multiple items are irrelevant, mismatched, or insufficiently grounded in the excerpt.\\
- \textbf{1}: Unsupported or hallucinated; anchors do not match the described content; evidence is largely invalid.\\

Evaluation Checklist (Evidence-First):\\
- Do the text snippets accurately reflect key points present in the excerpt?\\
- Are cited image anchors (\texttt{x1}, \texttt{x2}, \dots) relevant to their corresponding text snippets?\\
- Do the evidence items cover important aspects (methodology, results, conclusions) rather than superficial details?\\
- Is the evidence list well-organized and logically ordered (e.g., method $\rightarrow$ experiments $\rightarrow$ conclusions)?\\
- Does the judgment rely \textbf{only} on the provided excerpt (no external knowledge)?\\

Output Format (STRICTLY FOLLOW):\\
\texttt{Evidence Quality Score: <INTEGER 1-5>}\\
\texttt{Reason: <ONE PARAGRAPH JUSTIFICATION>}\\

Important Constraints:\\
- The score must be on its own line starting with \texttt{Evidence Quality Score:}.\\
- The score must be an integer from 1 to 5 (no decimals).\\
- Do not output any additional fields, bullet lists, or extra lines beyond the two required lines.\\
- Do not use Chinese characters or brackets like \texttt{\{ \}}.\\
- If an item cites an anchor that is not supported by the excerpt, you must penalize the score accordingly.\\

Example Output:\\
Evidence Quality Score: 4\\
Reason: The evidence items are mostly grounded in the excerpt: item 1 correctly cites x1 to describe the proposed architecture and matches the surrounding text, and item 2 references x2 for the main experimental comparison with accurate reported gains. However, item 3’s anchor-to-claim linkage is under-specified because the cited x3 is only loosely related to the stated ablation conclusion, reducing overall faithfulness despite generally coherent organization.
\end{tcolorbox}
\caption{Cross-validation prompt used to audit SIN-Summary evidence chains. It operationalizes an evidence-first rubric with discrete $1$--$5$ tiers and enforces a strict, machine-parsable two-line output schema, enabling scalable filtering of mismatched or weakly grounded image--text evidence.}
\label{fig:sin-summary-evidence-quality-prompt}
\end{figure*}

\nbf{(2) Cross-Validation.}
To systematically filter out plausible yet unverifiable instances, we employ a three-model jury system, with members dynamically sampled from Gemini-3-pro, GPT-5, Claude-4.5, and Qwen3-VL-Plus.
Each model independently audits candidate samples and assigns integer scores ranging from $1$ to $5$ across three dimensions: \textit{Question Validity} (assessing clarity, non-triviality, and document support), \textit{Answer Correctness} (ensuring factual consistency without critical hallucinations or omissions), and \textit{Evidence Consistency} (verifying relevance, sufficiency, and logical ordering).
To ensure high quality, we enforce a strict admission standard: we retain a sample only if it secures approval from the majority of the jury and achieves a minimum score of $4$ across every evaluation dimension.

The core objective of cross-validation is not to perform redundant generation but to transform subjective review into a scalable filtering mechanism using adjudicative, parsable, and evidence-centric prompts.
We uniformly adopt the following design principles across all validation prompts:
\begin{itemize}
\item \textbf{Evidence-First}: We explicitly require the review to center on evidence anchors and support relations, minimizing bias towards linguistic fluency or superficial plausibility.
\item \textbf{Criterionized Rubric}: We decompose the evaluation into fixed dimensions and discrete tiers ($1$--$5$) to reduce calibration drift between different auditor models.
\item \textbf{Strict I/O Schema}: We enforce fixed output fields and integer scores to facilitate automatic parsing and threshold-based filtering, avoiding the noise associated with free-form responses.
\item \textbf{Anchored Validation}: We explicitly verify the mapping between visual identifiers (e.g., $\langle x_k\rangle$) and text segments to detect image-text mismatches and irrelevant citations.
\item \textbf{Anti-Hallucination}: We require the judgment to rely strictly on support from the provided document segments and anchors, preventing completion based on external parametric knowledge.
\end{itemize}
As an illustrative example in~\cref{fig:sin-summary-evidence-quality-prompt}, the validation prompt for SIN-Summary requires the auditor to focus on ``Evidence Quality,'' checking whether summary items accurately cover key points of the paper, if citation anchors remain relevant, and if the organization follows a logical order.

\nbf{(3) Human-in-the-Loop Audit.}
Despite passing cross-validation, candidate samples may retain fine-grained grounding biases, such as anchors that are locatable but lack rigorous logical support, or the omission of key premises.
To address this, we assemble a review team consisting of $24$ graduate students and doctoral candidates.
Each sample undergoes independent verification by $2$--$3$ reviewers, while senior researchers adjudicate any disagreements.
This manual review strictly adheres to the following criteria:
\begin{itemize}
\item \textbf{Anchored Locatability}: Every $\langle x_k\rangle$ and text span must be precisely locatable within the SIN document, ensuring unambiguous citation.
\item \textbf{Rigor of Support}: The evidence chain must be sufficient and necessary to support the conclusion. For negative samples in SIN-Verify, the reviewer confirms that the attribute of insufficient or inconsistent holds true.
\item \textbf{Factual Consistency}: The answer or summary must not introduce information external to the document. Numerical values, comparative conclusions, and conditional settings must align strictly with the source text.
\item \textbf{Structural Compliance}: The output must conform to the task format and the specifications of the interleaved evidence chain, ensuring that the order of evidence matches the logical dependencies of the reasoning process.
\item \textbf{Discriminability}: Questions must avoid being overly trivial or undecidable. The evidence chain must possess diagnostic value, effectively distinguishing between evidence-driven reasoning'' and lucky guesses.''
\end{itemize}
This stage produces high-confidence golden samples and provides reliable signals for the update of seed examples.

\nbf{(4) Bootstrapping Iteration (Golden $\rightarrow$ Seed).}
The golden samples flow back into the system to serve as updated seed examples for the next cycle of synthesis and derivation.
This forms a closed feedback loop that continues until the benchmark meets the targets for scale and quality stability.

\nbf{Scale and Composition.}
Building upon a corpus of $4,000$ high-quality scientific documents, the multi-round synthesis pipeline yields approximately $3,200$ raw candidate samples.
Following the three-model cross-validation and Human-in-the-Loop review, \texttt{SIN-Bench} comprises a final set of $490$ golden samples: $159$ for SIN-Find, $158$ for SIN-QA, $89$ for SIN-Summary, and $84$ for SIN-Verify.

Furthermore, the volume of the currently released Golden Samples faces constraints due to the cost of human auditing. The number of high-quality samples is currently limited. 
However, the benchmark possesses inherent scalability due to the proposed semi-automated collaborative pipeline. 
Given sufficient computational and human resources, the pipeline encounters no bottleneck regarding the quantity of samples.

\section{On the Principle of ``No Evidence, No Score''}
\label{appx:no-evidence-no-score}

``No Evidence, No Score'' is an evidence-centered evaluation principle: a model should not obtain a high task score solely from answer-only guessing when its output cannot be grounded to verifiable document anchors. Accordingly, for evidence-required tasks, evidence quality is treated as a first-class component of the final score rather than an optional explanation.

Crucially, this principle targets \textit{verifiability} rather than citation surface forms. 
In practice, we do not penalize minor formatting variations (e.g., ``Fig.~2'' vs.\ ``Figure~2'' vs.\ ``x2'') as long as the anchor is recoverable and can be matched to the paper’s indexed visual markers. 
Only when the prediction provides no recoverable anchors (e.g., missing or invalid identifiers) does it receive zero evidence credit.

We also clarify how this principle remains logically self-consistent with separate answer scoring. 
We always compute AnsAcc as a standalone assessment of the predicted answer, but AnsAcc alone is not sufficient to yield a high overall task score without evidence. 
When no anchor is matched, evidence metrics collapse to zero by definition (e.g., $\mathrm{Matching}=0$, $\mathrm{F1}=0$; and we set $\mathrm{KT\text{-}sim}=0$ when it is undefined due to insufficient matches). 
This explicitly bounds the contribution of answer-only correctness and ensures that competitive performance requires grounded evidence, which is the intended behavior of ``No Evidence, No Score.''

Finally, our existing metrics already cover the practical corner cases that commonly arise in long-form scientific grounding. 
Over-generation of irrelevant anchors is reflected by reduced precision in Relevance; 
under-coverage of required evidence is reflected by reduced recall; 
and plausible but cherry-picked or mis-ordered evidence is captured by degradation in Logic. 
These behaviors are handled within the metric suite defined in the main text, without introducing additional indicators.

\begin{table}[!t]
\centering
\small
\resizebox{0.48\textwidth}{!}{
\begin{tabular}{lccc}
\toprule
\textbf{Task} & \textbf{Avg. Length} & \textbf{Max Length} & \textbf{Min Length} \\
\midrule
SIN-Find   & 108.89 & 186 & 60 \\
SIN-QA     & 122.77 & 203 & 78 \\
SIN-Verify & 243.49 & 322 & 160 \\
\bottomrule
\end{tabular}}
\caption{Token-length statistics of input contexts across task types.}
\label{tab:task_length_stats}
\end{table}

\begin{table}[!t]
\centering
\small
\resizebox{0.4\textwidth}{!}{
\begin{tabular}{lc}
\toprule
\textbf{Scope} & \textbf{Value} \\
\midrule
Tokens (Text / Image / Total) & 15k / 3k / 18k \\
Image Ratio & 15\% \\
Text Ratio    & 85\% \\
Avg. \#Images per instance    & 6.6 \\
\bottomrule
\end{tabular}}
\caption{Overall modality composition of benchmark inputs per sample.}
\label{tab:text_image_stats}
\end{table}

\begin{table}[!t]
\centering
\small
\resizebox{0.48\textwidth}{!}{
\begin{tabular}{lccc}
\toprule
\textbf{Task} & \textbf{Bold Avg.} & \textbf{Italic Avg.} & \textbf{Title Avg.} \\
\midrule
SIN-Find    & 70.52 & 122.46 & 14.64 \\
SIN-QA      & 70.83 & 118.17 & 15.03 \\
SIN-Summary & 37.46 & 155.74 & 14.91 \\
SIN-Verify  & 90.35 & 155.65 & 15.94 \\
\midrule
\textbf{Overall} & \textbf{68.01} & \textbf{132.81} & \textbf{15.03} \\
\bottomrule
\end{tabular}}
\caption{Markup statistics (bold/italic/title tokens) across tasks, reflecting structural formatting cues in the inputs.}
\label{tab:task_quality_features}
\end{table}

\subsection{General Statistics}

We report descriptive statistics of the benchmark inputs, focusing on (i) instance length, (ii) text–image composition, and (iii) structural markup density.

\cref{tab:task_length_stats} summarizes token-length statistics by task type. Overall, SIN-Verify instances are substantially longer than SIn-Find and SIN-QA (roughly $2$ times on average), indicating higher contextual and reasoning load, while SIN-Find and SIN-QA exhibit comparable length ranges.

\cref{tab:text_image_stats} characterizes the overall modality composition. On average, image-related tokens account for $15\%$ of the total tokens, with $6.6$ images per instance, suggesting that tasks require non-trivial visual grounding while remaining predominantly text-driven.

Finally, \cref{tab:task_quality_features} reports structural markup statistics (bold/italic/title tokens), which reflect the density of formatting cues and sectioning signals in the inputs. Such cues affect how models parse document structure and may influence both difficulty and computational cost across task types.

\begin{figure*}[!t] 
\centering
\includegraphics[width=1\textwidth]{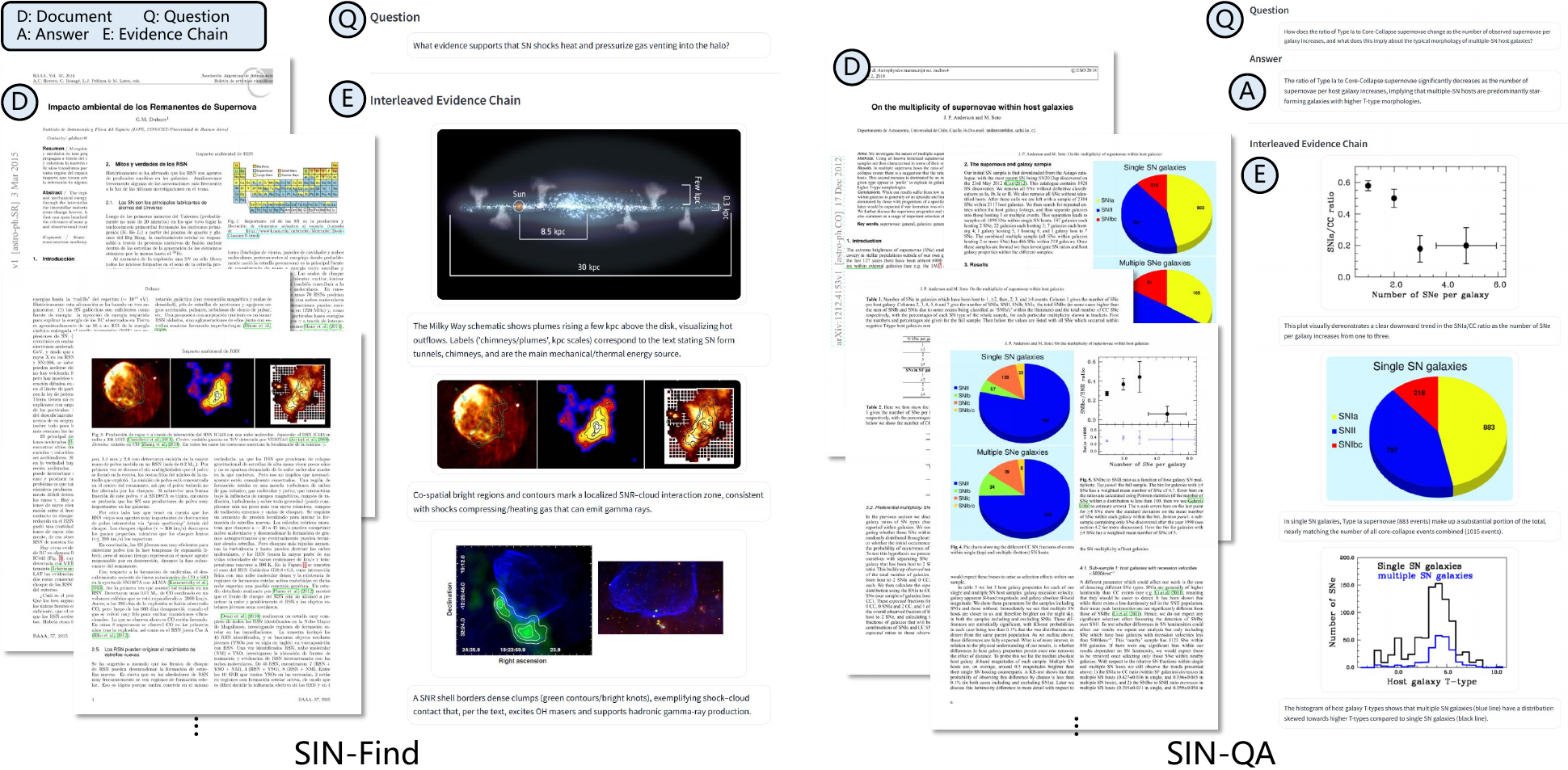}
\caption{Golden examples from \texttt{SIN-Bench}: SIN-Find and SIN-QA.
We visualize one curated instance per task with explicit markers for \textbf{D}ocument, \textbf{Q}uestion, \textbf{A}nswer, and the gold \textbf{E}vidence chain.}
\label{fig:gt-sample-find-qa} 
\end{figure*}

\begin{figure*}[!t] 
\centering
\includegraphics[width=1\textwidth]{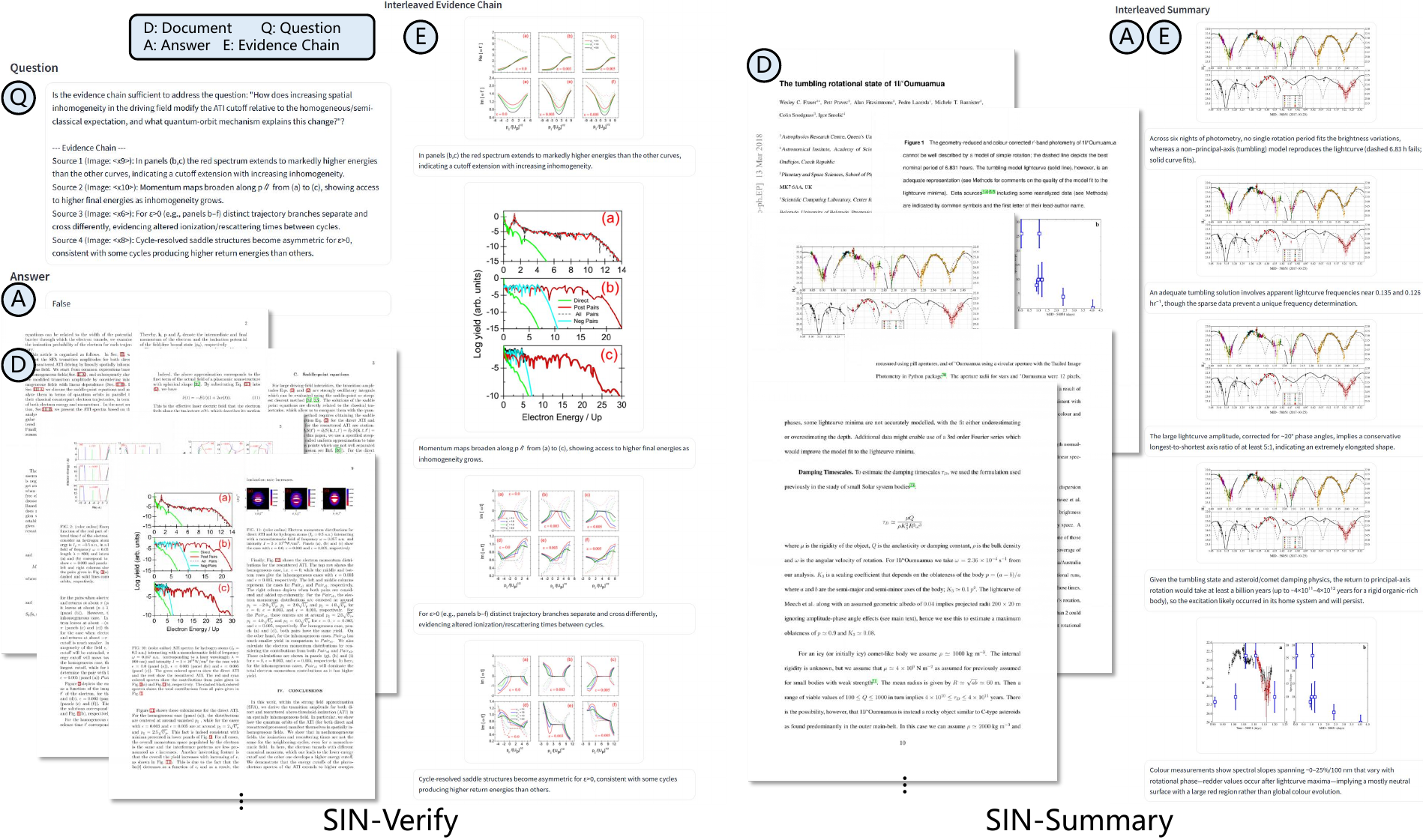}
\caption{Golden examples from \texttt{SIN-Bench}: SIN-Verify and SIN-Summary.
We visualize one curated instance per task with explicit markers for \textbf{D}ocument, \textbf{Q}uestion, \textbf{A}nswer, and the gold \textbf{E}vidence chain.}
\label{fig:gt-sample-ver-sum} 
\end{figure*}

\subsection{Samples from \texttt{SIN-Bench}}

As shown in~\cref{fig:gt-sample-find-qa} and~\cref{fig:gt-sample-ver-sum}, we visualize one ``golden'' instance per task in \texttt{SIN-Bench}. 
Since scientific documents are long, rendering full instances would be space-inefficient and would dilute the task-relevant signal. 
We therefore show only cropped screenshots that cover the document, the question, the annotated evidence region(s), and the gold output (answer, verification decision, or summary items). 
The benchmark itself operates on the full interleaved source files (Markdown with image assets), which are released with stable identifiers for exact reconstruction.

\subsection{License}

The benchmark dataset and source code are available under the MIT License. 
The full text of the license is provided at~\url{https://opensource.org/licenses/MIT}.

\subsection{Discipline Taxonomy List}
\label{ass:discipline-taxonomy}

To ensure a comprehensive evaluation of multimodal reasoning across scientific disciplines, we constructed a three-level hierarchical taxonomy referencing the arXiv subject classification. 

Our pipeline integrates large model capabilities with expert human verification. 
We first utilized Qwen3-VL-2B to perform preliminary classification on the collected samples. 
Subsequently, domain experts reviewed the results to correct misclassifications and refine subfield definitions.

The resulting taxonomy comprises $12$ Level-1 major disciplines, branching into $35$ Level-2 domains and $84$ Level-3 subfields. The hierarchical structure is organized as follows:

\begin{itemize}[nosep,leftmargin=*]

  \item \textbf{Astronomy \& Astrophysics}:
  \begin{itemize}[nosep,leftmargin=*]
    \item Extragalactic Astronomy \& Cosmology: Active Galactic Nuclei \& Quasars; Galaxy Formation \& Evolution.
    \item Galaxy Evolution \& Structure: Structure \& Evolution of Galaxies.
    \item Planetary \& Space Science: Planetary Geology \& Exploration.
    \item Planetary Systems \& Formation: Planetary Systems Protoplanetary Disks.
  \end{itemize}

  \item \textbf{Biology \& Life Sciences}:
  \begin{itemize}[nosep,leftmargin=*]
    \item Neuroscience: Neuroimaging.
  \end{itemize}

  \item \textbf{Computer Science}:
  \begin{itemize}[nosep,leftmargin=*]
    \item Algorithms \& Theory: Combinatorial Algorithms; Graph \& Combinatorial Algorithms; Optimization; Quantum Algorithms \& Computation Theory.
    \item Communication \& Networking: Information Theory \& Coding; Channel Modeling.
    \item Computer Vision: 2D 3D Reconstruction \& Modeling; Image Processing \& Restoration; Video Understanding \& Compression; Vision.
    \item Electrical \& Computer Engineering: Embedded \& Hardware Systems.
    \item Human-Computer Interaction: Virtual Augmented Reality \& Haptics.
    \item Machine Learning \& AI: Active Learning; Clustering \& Unsupervised Learning; Deep Learning; Deep Reinforcement Learning; Differential Privacy \& Anonymization; Explainable \& Fair AI; Federated \& Continual Learning; Generative Models; Graph Neural Networks; Multimodal Learning \& Fusion; Nonlinear Time Series; Object Detection; Particle Filters \& Data Assimilation; Privacy; Quantum Algorithms \& Computation Theory; Second Order Methods; Submodular Optimization.
    \item Natural Language Processing: Computational Linguistics; Information Retrieval \& QA Systems.
    \item Robotics: Autonomous \& Mobile Robotics; Learning for Robotic Control.
    \item Security \& Privacy: Adversarial Machine Learning; Blockchain \& Distributed Ledger Security; Differential Privacy \& Anonymization.
    \item Software Engineering \& Programming: AI for Software Engineering; Program Synthesis \& Verification.
    \item Systems \& Infrastructure: Edge \& Cloud Computing \& Serverless.
  \end{itemize}

  \item \textbf{Earth \& Environmental Sciences}:
  \begin{itemize}[nosep,leftmargin=*]
    \item Climate \& Atmospheric Science: Atmospheric Chemistry \& Air Quality; Climate Modeling \& Prediction.
    \item Geophysics \& Remote Sensing: Geospatial Intelligence \& GIS; Satellite \& Aerial Imaging.
    \item Oceanography \& Hydrology: Water Resources \& Hydrology.
    \item Planetary \& Space Science: Planetary Geology \& Exploration; Space Instrumentation \& Missions; Space Weather \& Near Earth Environment.
  \end{itemize}

  \item \textbf{Economics \& Finance}:
  \begin{itemize}[nosep,leftmargin=*]
    \item Macroeconomics \& International Trade: Sovereign Debt \& Default Risk.
  \end{itemize}

  \item \textbf{Engineering}:
  \begin{itemize}[nosep,leftmargin=*]
    \item Mechanical \& Robotics Engineering: Robotics \& Automation.
  \end{itemize}

  \item \textbf{Mathematics}:
  \begin{itemize}[nosep,leftmargin=*]
    \item Pure Mathematics: Analysis \& Functional Analysis.
  \end{itemize}

  \item \textbf{Medicine \& Health Sciences}:
  \begin{itemize}[nosep,leftmargin=*]
    \item Neurology \& Psychiatry: Computational Psychiatry.
  \end{itemize}

  \item \textbf{Philosophy \& Ethics}:
  \begin{itemize}[nosep,leftmargin=*]
    \item AI \& Technology Ethics: Philosophy of Quantum Physics.
  \end{itemize}

  \item \textbf{Physics}:
  \begin{itemize}[nosep,leftmargin=*]
    \item Astrophysics \& Cosmology: Astronomical Instrumentation; Atmospheric Chemistry \& Air Quality; Cosmology \& Galaxy Formation; Exoplanets \& Planetary Science; Gravitational Lensing \& Cosmology; Gravitational Lensing \& Microlensing; Neutron Stars \& Pulsars; Planetary Geology \& Exploration; Satellite \& Aerial Imaging; Space Instrumentation \& Missions; Supernova Remnants \& Cosmic Rays; Supernovae \& Stellar Evolution.
    \item Condensed Matter \& Materials: Computational Materials Physics; Condensed Matter Physics; Materials Science \& Characterization; Spin Glass \& Magnetic Materials; Superconductivity \& Magnetism.
    \item Optics, Fluids \& Dynamics: Laser Physics \& Photonics; Photonic Crystals \& Band Gap Physics; Plasma \& Fusion Physics; Quantum Optics \& Atomic Physics .
    \item Optics \& Photonics: Nonlinear Optics \& Optical Processing; Optical Imaging \& Sensing. 
    \item Quantum \& Particle Physics: Gravitational Waves \& Cosmology; High; Neutrino Astronomy \& Astrophysics; Quantum Computing \& Information; Quantum Optics \& Atomic Physics.
  \end{itemize}

  \item \textbf{Social Sciences}:
  \begin{itemize}[nosep,leftmargin=*]
    \item Communication, Media \& Linguistics: Computational Social Media Analysis.
    \item Economics \& Finance: Econometrics \& Applied Microeconomics.
    \item Education \& Learning Sciences: Educational Technology \& AI in Education.
    \item Sociology \& Social Policy: Social Networks \& Computational Social Science.
  \end{itemize}

  \item \textbf{Statistics \& Data Science}:
  \begin{itemize}[nosep,leftmargin=*]
    \item Data Science Applications: Health \& Social Data Science.
    \item Statistical Theory \& Methods: Nonparametric \& Bayesian Statistics.
  \end{itemize}

\end{itemize}

This structured distribution ensures that \texttt{SIN-Bench} covers a wide spectrum of knowledge.

\section{Comparisons with Existing Benchmarks}

\begin{table*}[!t]
\centering
\small
\resizebox{\textwidth}{!}{
\setlength{\tabcolsep}{4pt}
\renewcommand{\arraystretch}{1.15}
\begin{tabular}{p{3.3cm} p{2.5cm} p{2.2cm} p{2.0cm} p{3.2cm} p{3.8cm} cc}
\toprule
\textbf{Benchmark} & \textbf{Doc type} & \textbf{Scale} & \textbf{Avg. length} & \textbf{Domain coverage} & \textbf{Main tasks} & \textbf{Ev.} & \textbf{Ev.-score} \\
\midrule
\makecell[l]{MMLongBench-Doc\\ \citep{DBLP:conf/nips/MaZC0JLLLMDZP0W24@mmlongbench-doc}}
& \makecell[l]{Scientific PDFs\\ (multi-domain)}
& \makecell[l]{130--135 docs\\ (benchmark)}
& \makecell[l]{ \(\sim\)21k tokens}
& \makecell[l]{7 domains\\ (e.g., acad./legal/tech)}
& \makecell[l]{Long-doc understanding;\\ numerical reasoning;\\ cross-element search}
& No & No \\
\midrule
\makecell[l]{DocGenome (test)\\ \citep{DBLP:journals/corr/abs-2406-11633@DocGenome}}
& \makecell[l]{Scientific papers\\ (structured doc)}
& \makecell[l]{--}
& \makecell[l]{--}
& \makecell[l]{8 classes\\ +153 subclasses}
& \makecell[l]{7 doc tasks (e.g.,\\ cls./grounding/QA/\\ layout detection, \dots)}
& No & No \\
\midrule
\makecell[l]{Document Haystack\\ \citep{DBLP:journals/corr/abs-2507-15882@document-haystack}}
& \makecell[l]{Long PDFs\\ (general)}
& \makecell[l]{400 docs}
& \makecell[l]{5--200 pp.}
& \makecell[l]{General}
& \makecell[l]{Needle-in-a-haystack\\ retrieval (long-doc)}
& No & No \\
\midrule
\makecell[l]{SciFIBench\\ \citep{DBLP:conf/nips/00040HA24@SciFIBench}}
& \makecell[l]{Scientific figures\\ (charts)}
& \makecell[l]{--}
& \makecell[l]{--}
& \makecell[l]{8 figure types\\ (CS-centric)}
& \makecell[l]{Figure interpretation}
& No & No \\
\midrule
\rowcolor{promptbg}
\textbf{SIN-Bench (ours)}
& \makecell[l]{Scientific docs\\ (interleaved)}
& \makecell[l]{231 docs\\ / 490 inst.}
& \makecell[l]{\(\sim\)18k tokens}
& \makecell[l]{10+ L1 domains;\\ 30+ L2; 80+ L3}
& \makecell[l]{4 tasks: Find/Verify/\\ QA/Summary\\ (discovery\(\to\)synthesis)}
& \textbf{Yes} & \textbf{Yes} \\
\bottomrule
\end{tabular}}
\caption{Comparison with representative long-document and scientific multimodal benchmarks. 
``Ev.'' indicates whether the benchmark requires an explicit evidence chain in the model output, and ``Ev.-score'' indicates whether evidence quality is explicitly evaluated (beyond answer-only scoring). }
\label{tab:benchmark_comparison}
\end{table*}

\cref{tab:benchmark_comparison} positions \texttt{SIN-Bench} against representative long-document and scientific multimodal benchmarks. 
Prior benchmarks have substantially advanced \emph{long-context stress testing} and \emph{document-centric task coverage}, including long-PDF understanding suites (e.g., MMLongBench-Doc) and needle-style retrieval tests that insert sparse targets into lengthy contexts (e.g., Document Haystack).
Scientific-domain resources further expand coverage via structured document tasks (DocGenome (test)) or figure-focused evaluations (SciFIBench).
Despite these advances, most existing benchmarks still rely primarily on answer-level grading and do not explicitly require models to produce auditable cross-modal evidence chains, which makes it difficult to disentangle document-grounded reasoning from plausible but ungrounded generation.

In contrast, \texttt{SIN-Bench} targets evidence-based scientific comprehension over native scientific documents in an interleaved image and text format, and organizes four progressive tasks (Find/Verify/QA/Summary) that mirror a discovery$\rightarrow$verification$\rightarrow$synthesis workflow. 
Beyond answer correctness, we explicitly score evidence quality (e.g., Matching/Relevance/Logic), enabling fine-grained diagnosis of where failures arise (missing prerequisites vs. spurious citations) rather than collapsing them into an answer-only signal. 
While \texttt{SIN-Bench} contains fewer evaluated documents than some large-scale resources, it is distilled from an initial pool of $50$k collected papers through a multi-stage cleaning and filtering pipeline designed to preserve high-quality scientific documents at each step. 
Importantly, this pipeline is \textit{scalable}: human involvement is lightweight and predominantly verification-based (binary/ordinal judgments), avoiding the need for extensive manual authoring of questions and reference answers while still producing reliable, document-grounded evaluation instances.

\section{Details of Experimental Setting}
\label{as:details-of-exp-set}

\subsection{Baselines}
\label{ass:details-baselines}

We report the version identifiers for all proprietary models evaluated to support reproducibility. 
We run all evaluations once with the temperature set to $0$, including open-weight models. 
The evaluated proprietary models and their version identifiers are as follows:
\begin{itemize}
\item Gemini-3-pro: \texttt{gemini-3-pro-preview-11-2025}
\item Gemini-2.5-pro: \texttt{gemini-2.5-pro-thinking}
\item Claude-sonnet-4.5: \texttt{claude-sonnet-4-5-20250929-thinking}
\item GPT-5: \texttt{gpt-5}
\item Grok-4: \texttt{grok-4-0709}
\end{itemize}

\subsection{Models Used in the Benchmark and Experiments}
\label{ass:model-in-paper}

We evaluate a mix of proprietary and open-source models in \texttt{SIN-Bench}. 
In the analysis experiments, compute budget and iteration cost constrain the number of controlled ablations, and the experimental design therefore prioritizes a representative strong model as the primary testbed. 
Gemini-3-pro serves as this primary model since it achieves the best overall performance on \texttt{SIN-Bench} when averaging across tasks, which makes it a representative reference point for diagnostic analyses.

In the modality study in~\cref{fig:interleaved_analysis} (\cref{ss:discussion}), we isolate the impact of raw visual inputs by constructing a ``pure text'' setting.
This setting replaces each image slot in the original interleaved sequence with a natural-language description of the corresponding figure. 
The descriptions are generated by Qwen3-VL-8B-Instruct, which preserves the document structure while removing direct visual signals.

For evaluation components that require semantic judgment, we adopt an LLM-based evaluator instead of n-gram overlap metrics.
This design reduces sensitivity to paraphrase and surface-form variation, which n-gram metrics often penalize despite semantic equivalence.
It also fits settings with diverse valid answer phrasings, where a single reference string can under-specify correctness and cause systematic false negatives.
In addition, an LLM judge supports rubric-guided, criteria-based grading that is directly aligned with task definitions.
Prior work shows that rubric- or form-based LLM evaluation improves agreement with human judgments and provides a scalable assessment for open-ended outputs~\citep{DBLP:conf/emnlp/LiuIXWXZ23@g-eval,DBLP:conf/nips/ZhengC00WZL0LXZ23@mt-bench}.
Concretely, Matching and answer correctness (AnsAcc) use Qwen3-8B as the automatic judge.
Furthermore, in~\cref{ass:judge-in-metrics}, we benchmark the judge against human expert judgments and observe close agreement, which supports the reliability of the LLM-based evaluator for scalable evaluation.

\subsection{Judges in Metrics}
\label{ass:judge-in-metrics}

\begin{table}[!t]
\centering
\small
\resizebox{0.48\textwidth}{!}{
\begin{tabular}{llcc}
\toprule
\textbf{Task} & \textbf{Metric} & \textbf{Pearson ($r$)} & \textbf{Spearman ($\rho$)} \\
\midrule
SIN-Find      & Matching (M)       & 0.824 & 0.795 \\
\midrule
\multirow{2}{*}{SIN-QA} & Matching (M) & 0.812 & 0.788 \\
                        & Answer Acc. & 0.895 & 0.862 \\
\midrule
SIN-Summary       & Matching (M)       & 0.768 & 0.741 \\
\midrule
- & \textbf{Average}            & \textbf{0.825} & \textbf{0.797} \\
\bottomrule
\end{tabular}}
\caption{Correlation between human expert ratings and automated scores (evaluated by Qwen3-8B) across different tasks and metrics. All results are statistically significant with $p < 0.001$.}
\label{tab:consistency_results}
\end{table}

To ensure the reliability of our automated evaluation in matching and question accuracy metrics, we employ Qwen3-8B as the judge.
We validated this choice through a rigorous consistency study involving all instances for each task.
For these samples, we applied a dual-scoring protocol to compare the automated judge against human experts.
Specifically, Qwen3-8B~\citep{DBLP:journals/corr/abs-2505-09388@qwen3} independently evaluated the model predictions, while a panel of $2$--$3$ Master's and Ph.D. students with relevant domain expertise performed manual evaluation.
The human ratings were aggregated via majority voting to establish the evaluation gold standard.
We then computed the correlation between the automated scores and this human consensus.
As shown in~\cref{tab:consistency_results}, the automated judge exhibits strong alignment with the gold standard (Avg Pearson $r=0.825$), confirming that Qwen3-8B serves as a valid and consistent surrogate for human evaluation.

\section{Additional Experiments and Result Analysis}
\label{as:add-experiments-analysis}

\begin{figure}[!t] 
\centering
\includegraphics[width=0.48\textwidth]{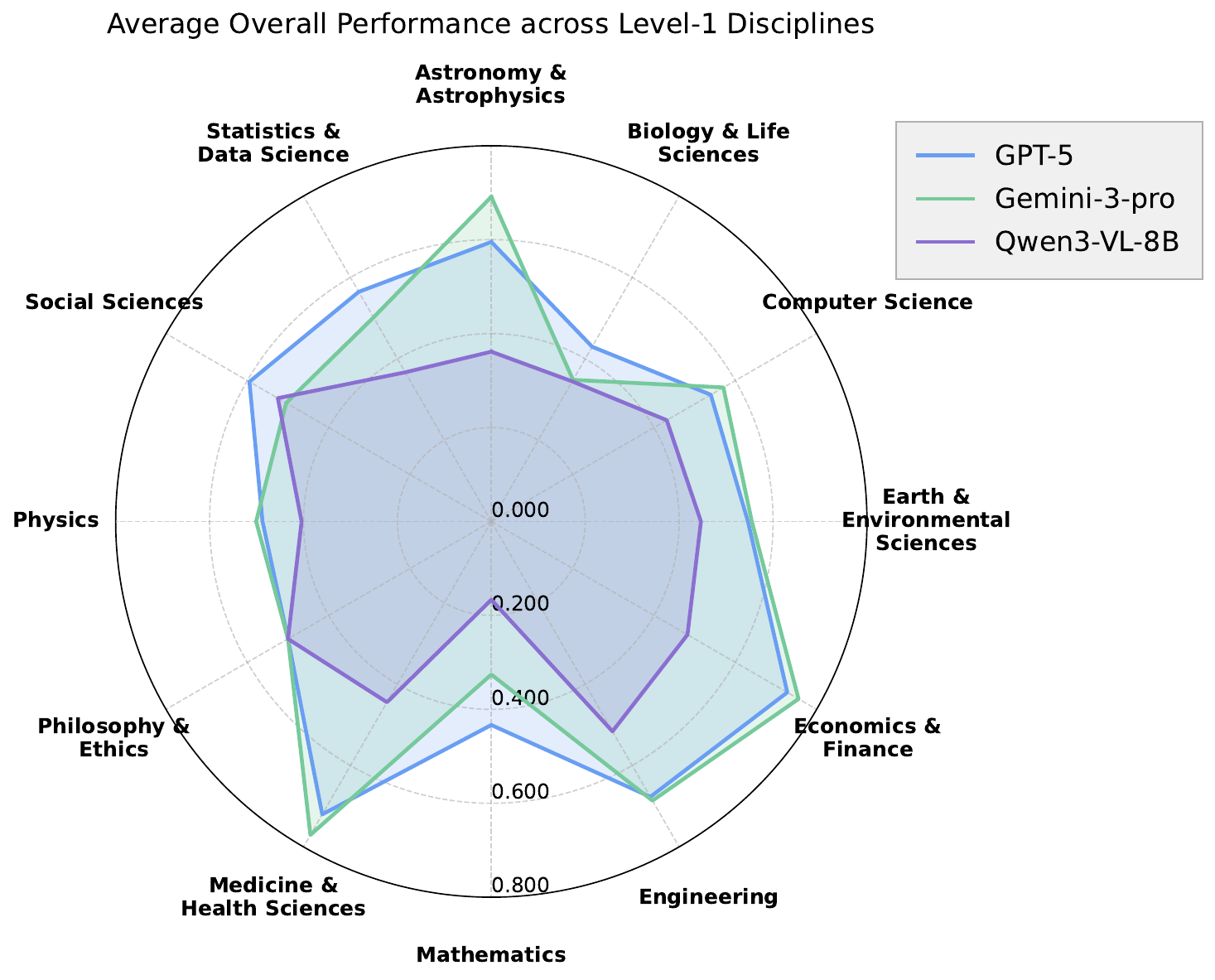} 
\caption{Domain-wise performance on \texttt{SIN-Bench}. Radar plot of average overall scores (across all tasks) for GPT-5, Gemini-3-pro, and Qwen3-VL-8B across $12$ primary disciplines. Higher is better, highlighting substantial domain-dependent variability in evidence-grounded scientific comprehension.}
\label{fig:domain-result} 
\end{figure}

\subsection{Discipline-level Performance Analysis}
\label{ass:discipline-analysis}

\cref{fig:domain-result} summarizes discipline-level performance under our Level-1 taxonomy. 
We observe the cross-discipline variation, indicating that \texttt{SIN-Bench} captures meaningful shifts in evidence-grounded scientific comprehension. 
Aggregated across models, \textit{Economics \& Finance} and \textit{Medicine \& Health Sciences} exhibit relatively strong overall performance (avg.\ $\approx0.66$ and $0.65$), whereas \textit{Mathematics} is consistently the most demanding domain (avg.\ $\approx0.31$). 
This contrast suggests that rigorous symbolic manipulation and quantitative grounding remain key bottlenecks for current MLLMs.

Model rankings are also discipline-dependent. 
Gemini-3-pro shows clear advantages in visually intensive or layout-heavy disciplines such as \textit{Astronomy \& Astrophysics} and \textit{Medicine \& Health Sciences} (e.g., $+0.01$ and $+0.05$ over GPT-5), consistent with stronger cross-modal anchoring when evidence is primarily conveyed through figures, tables, and chart-like graphics. 
In contrast, GPT-5 performs better in more formal/quantitative disciplines, most notably \textit{Mathematics} ($+0.10$ over Gemini-3-pro) and \textit{Statistics \& Data Science}. 
Qwen3-VL-8B trails the frontier models in most disciplines, yet remains comparatively competitive in \textit{Social Sciences} and converges with the others in \textit{Philosophy \& Ethics}, where the decision signal may rely more on general discourse understanding and less on fine-grained multimodal evidence.

\begin{figure}[!t] 
\centering
\includegraphics[width=0.48\textwidth]{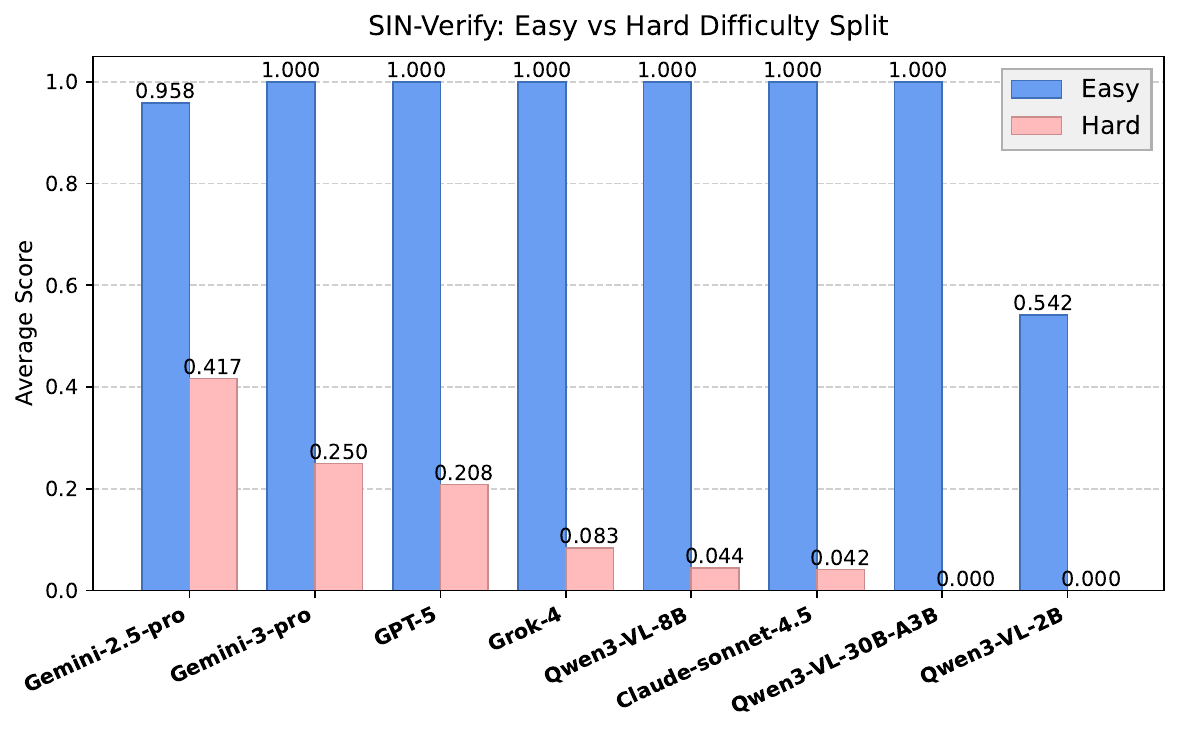}
\caption{Accuracy on SIN-Verify with two negative sets. Easy negatives are clearly irrelevant evidence; hard negatives are cross-validation near-misses with correct answers but insufficient/ambiguous evidence support.}
\label{fig:verify-hard-easy-full-result} 
\end{figure}

\begin{figure*}[!t] 
\centering
\includegraphics[width=0.8\textwidth]{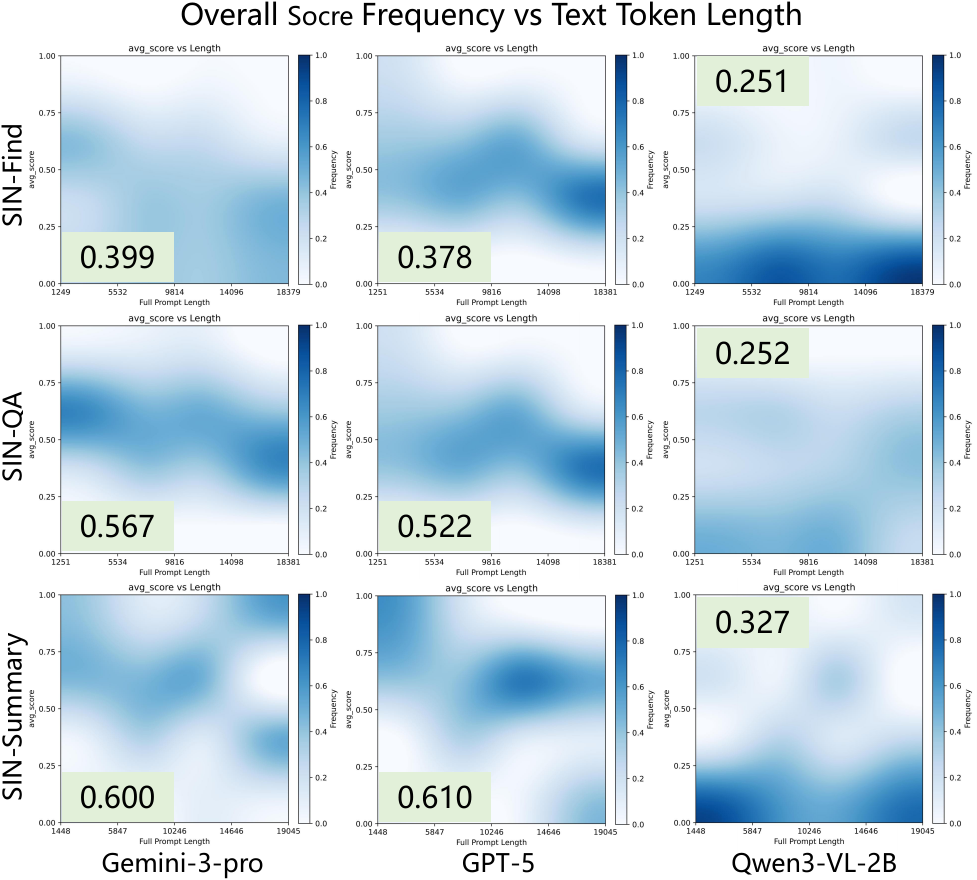}
\caption{Overall score frequency vs. text-token length for SIN-Find, SIN-QA and SIN-Summary. Normalized 2D frequency of samples binned by text prompt length and overall scores.}
\label{fig:text_len}
\end{figure*}

\begin{figure*}[!t] 
\centering
\includegraphics[width=0.8\textwidth]{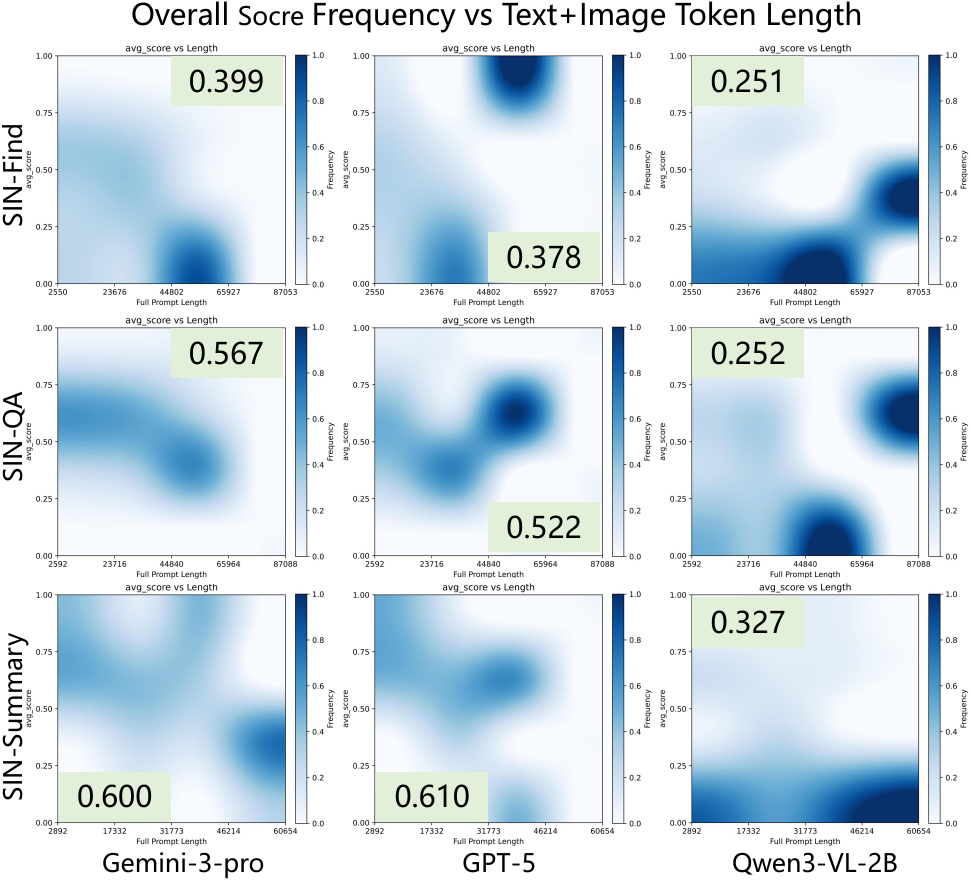}
\caption{Overall score frequency vs. total (text and image) token length. Same as \cref{fig:text_len} but with prompt length computed from text plus image tokens from the API request.}
\label{fig:total_len}
\end{figure*}

\subsection{An Analysis of Near-Miss Evidence in SIN-Verify}
\label{ass:sin-verify-hard-easy}

To probe verification difficulty beyond trivial mismatch, we construct two negative sets during cross-validation (\cref{ass:sin-bench-pipeline}, step (2)). 
Easy negatives pair a question with evidence that is clearly irrelevant, making rejection straightforward. 
Hard negatives are near-miss cases: the question quality is high, and the answer is correct, but the evidence chain fails cross-validation agreement, i.e., it is ambiguous or insufficient to conclusively support the answer. 
We sample $24$ instances for each setting.

As shown in~\cref{fig:verify-hard-easy-full-result}, MLLMs nearly saturate on easy negatives, indicating strong coarse mismatch detection. 
In contrast, accuracy drops sharply on hard negatives, indicating that near-miss evidence makes verification substantially more demanding than relevance detection. 
Models must assess evidence sufficiency via fine-grained cross-modal grounding (e.g., numbers, conditions, and figure–text alignment) while avoiding answer-driven rationalization.
Gemini-2.5-pro performs best on hard negatives ($0.417$), consistent with stronger calibration toward ``insufficient evidence'' decisions. 
Gemini-3-pro ($0.250$) and GPT-5 ($0.208$) remain substantially below the ceiling, implying that general reasoning strength does not directly translate to stricter support judgments under near-miss evidence.
The remaining models fall to near-zero, aligning with verification behavior dominated by surface relevance cues.

\subsection{Impact of Document Length on Performance Stability.}
\label{ass:impact-doc-len}

Scientific papers are long-form and multimodal by construction. To characterize how performance varies with context length under interleaved inputs, we analyze the joint distribution between prompt length and overall score using two token accounting schemes: text-only tokens (\cref{fig:text_len}) and text+image tokens (\cref{fig:total_len}). 
In this subsection, we report results for SIN-Find, SIN-QA, and SIN-Summary. 
In~\cref{ss:discussion}, we further focus the interpretation on SIN-QA and SIN-Summary, since these two tasks are high-level and require holistic document understanding plus evidence-backed generation.

\nbf{Setup.}
For each evaluated sample, we record the true input token count used in the API call and the sample's overall score. 
In~\cref{fig:text_len}, the x-axis counts text tokens only. 
We adopt this view as the primary fairness axis because English tokenization is broadly similar across models, while image tokenization can be strongly model- and pipeline-dependent. 
Therefore, we consider text-only token length a more representative and fair basis for cross-model comparison.
In \cref{fig:total_len}, the x-axis counts text$+$image tokens, where image tokens depend on visual encoding policies (e.g., native-resolution v.s. resized inputs). 
For Qwen3-VL-2B, token statistics follow the official configuration. 
Each panel visualizes the normalized frequency over (prompt length, overall score); the inset reports the mean overall score.

\nbf{Text-only length (\cref{fig:text_len}): robustness under long textual contexts.}
Across all three tasks, Gemini-3-pro and GPT-5 place substantial probability mass in the mid-to-high score region across the full text-length range (roughly $1$k--$19$k tokens). 
GPT-5 shows a more pronounced score drop than Gemini-3-pro as prompts become longer. 
However, neither model exhibits a systematic collapse of probability mass into the low-score band at the longest contexts. 
This pattern suggests that, for strong models, increased text length alone is unlikely to be the primary bottleneck; residual errors more plausibly stem from selective evidence identification and fine-grained grounding in dense scientific narratives.

In contrast, Qwen3-VL-2B exhibits a consistently low-scoring regime. 
For SIN-Find and SIN-QA, the score distribution concentrates near the lower end, with limited migration toward higher scores even at shorter text lengths. 
A similar skew is observed for SIN-Summary, where performance remains dominated by low-score mass throughout. 
Overall, these trends indicate that the limiting factor for Qwen3-VL-2B is not merely long-context handling, but a broader gap in robust evidence retrieval, cross-modal alignment, and high-level synthesis under interleaved scientific inputs.

\nbf{Text$+$image length (\cref{fig:total_len}): length expansion and clustering induced by visual tokenization.}
When image tokens are included, the prompt-length range expands substantially (up to $\sim$87k tokens). 
This expansion largely reflects figure count and model-specific visual encoding. 
Relative to~\cref{fig:text_len}, the distributions also become more clustered along the x-axis. 
This is consistent with total length discretizing into regimes induced by typical figure counts/resolutions and the underlying visual tokenizer. 
Therefore, trends along the total-length axis should be interpreted cautiously, as it conflates document visual content with preprocessing and encoding choices.

\begin{figure*}[!t] 
\centering
\includegraphics[width=1\textwidth]{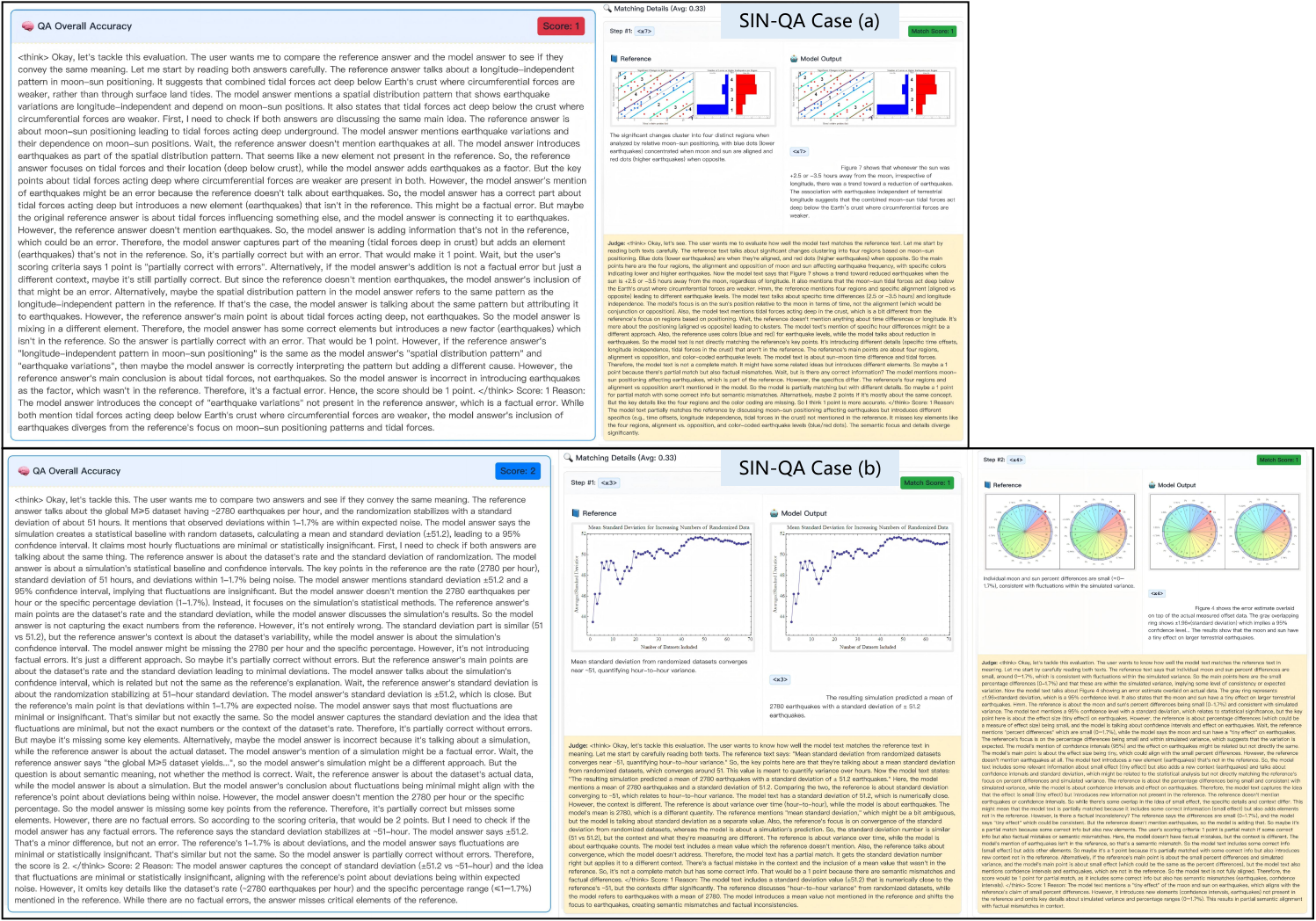}
\caption{Error cases on SIN-QA.}
\label{fig:more-cases}
\end{figure*}

Gemini-3-pro and GPT-5 attain similar mean overall scores, but their distributions differ qualitatively as context grows. 
Gemini-3-pro remains comparatively smooth and spread across mid-to-high scores, with a consistent trend across lengths. 
By contrast, GPT-5 exhibits a clearer bifurcation: as total length increases, probability mass splits, with one mode staying in high-score regions while another shifts toward lower scores. 
A possible explanation is that total length mainly increases through additional figures, which amplifies variance in visual grounding difficulty. 
When the added figures are redundant or directly supportive, GPT-5 can exploit the extra multimodal evidence and retain high scores. 
However, when longer inputs introduce visually similar distractors, dense multi-figure evidence chains, or resolution-sensitive cues (e.g., small text, numbers, axis labels), errors become more likely, yielding a low-score mode. 
In other words, longer multimodal contexts do not uniformly degrade performance; instead, they increase instance-level variance by mixing ``helpful'' and ``adversarial'' visual additions.

For Qwen3-VL-2B, the score distribution is still dominated by low-score regions across tasks. 
Nevertheless, in SIN-Find and SIN-QA, we observe non-trivial mass at moderate scores under longer total lengths. 
One likely reason is that these longer prompts are driven more by additional images than by substantially longer text, and the tasks can sometimes be solved via salient visual cues or near-surface matching once the correct figure is present. 
In contrast, SIN-Summary remains almost entirely concentrated in low-score regions even at large total lengths, suggesting that additional visual tokens alone do not alleviate the core difficulty of globally organizing and synthesizing evidence into a coherent, document-level summary.

\section{Case Study}
\label{as:case-study}

As shown in~\cref{fig:more-cases}, we further analyze SIN-QA, since it requires both a final answer and an explicit evidence chain, making it suitable for diagnosing grounding failures beyond aggregate scores. 
In addition to the two failure modes summarized in~\cref{ss:discussion} (Information Deficiency and Spurious Reasoning), our case studies reveal two specific instantiations.

In Case (a), the model injects unsupported concepts absent from the reference (e.g., expanding a tidal-force explanation into an earthquake narrative). While some statements remain superficially related, the added content shifts the explanatory frame and induces semantic drift, exemplifying \textbf{Spurious Reasoning}: fluent and confident outputs contaminated by ungrounded expansions that reduce precision. 
In contrast, Case (b) reflects a procedural form of \textbf{Information Deficiency}: 
the model does not adhere to the reference's intended statistical interpretation, by conflating or oversimplifying the role of standard deviation in variability analysis. 
The resulting answer appears quantitative but violates key methodological constraints, leading to method-inconsistent conclusions.

\end{document}